\documentclass[journal]{IEEEtran}
\ifCLASSINFOpdf
\else
\fi
\usepackage{graphicx}
\usepackage{amsfonts}       
\usepackage{amsmath}
\usepackage{amsthm}
\usepackage{soul}
\usepackage{color}

\usepackage{hyperref}       
\usepackage{url}            
\usepackage{booktabs}       
\usepackage{nicefrac}       
\usepackage{microtype}      
\usepackage{comment}
\usepackage{bm}
\usepackage{multirow}
\usepackage{enumerate}
\usepackage{lineno}
\usepackage{makecell}

\usepackage{algorithm}
\usepackage[noend]{algpseudocode}

\newtheorem{definition}{Definition}
\newtheorem{theorem}{Theorem}

\newcommand{\bi}{\begin{itemize}}
\newcommand{\ei}{\end{itemize}}
\newcommand{\ba}{\begin{array}}
\newcommand{\ea}{\end{array}}

\DeclareMathOperator*{\argmax}{arg\,max}

\usepackage{xspace}
\newcommand{\model}{{TGV-CRN}\xspace}
\newcommand{\modelnospace}{{TGV-CRN}}

\newtheorem{assumption}{Assumption}[section]
\newcommand{\indep}{\perp\!\!\!\!\perp} 

\newcommand{\eq}[1]{\begin{align}#1\end{align}}

\newcommand{\bmx}[0]{\begin{bmatrix}}
\newcommand{\emx}[0]{\end{bmatrix}}

\begin{document}
%
\title{Estimating counterfactual treatment outcomes over time in complex multiagent scenarios}
%
%
%

\author{Keisuke Fujii,~\IEEEmembership{Member,~IEEE,}
        Koh Takeuchi, 
        Atsushi Kuribayashi,
        Naoya Takeishi,\\
        Yoshinobu Kawahara,
        and~Kazuya Takeda,~\IEEEmembership{Senior~Member,~IEEE}
\thanks{K. Fujii is with Graduate School of Informatics at Nagoya University, JAPAN and RIKEN Center for Advanced Intelligence Project, JAPAN. e-mail: fujii@i.nagoya-u.ac.jp.}
\thanks{K. Takeuchi is with Graduate School of Informatics at Kyoto University and RIKEN.}
\thanks{A. Kuribayashi and K. Takeda are with Graduate School of Informatics at Nagoya University.}
\thanks{N. Takeishi is with Graduate School of Engineering at the University of Tokyo and RIKEN Center for Advanced Intelligence Project.}
\thanks{Y. Kawahara is with Graduate School of Information Science and Technology at Osaka University and RIKEN Center for Advanced Intelligence Project.}}

%
%

\markboth{Submitted to IEEE TNNLS, Special Issue: Information Theoretic Methods for the Generalization, Robustness and Interpretability...}
{Fujii \MakeLowercase{\textit{et al.}}: Estimating counterfactual treatment outcomes over time in multiagent}
%



\maketitle

\begin{abstract}
Evaluation of intervention in a multiagent system, e.g., when humans should intervene in autonomous driving systems and when a player should pass to teammates for a good shot, is challenging in various engineering and scientific fields. 
Estimating the individual treatment effect (ITE) using counterfactual long-term prediction is practical to evaluate such interventions.
However, most of the conventional frameworks did not consider the time-varying complex structure of multiagent relationships and covariate counterfactual prediction. 
This may lead to erroneous assessments of ITE and difficulty in interpretation.
Here we propose an interpretable, counterfactual recurrent network in multiagent systems to estimate the effect of the intervention.
Our model leverages graph variational recurrent neural networks and theory-based computation with domain knowledge for the ITE estimation framework based on long-term prediction of multiagent covariates and outcomes, which can confirm the circumstances under which the intervention is effective.   
On simulated models of an automated vehicle and biological agents with time-varying confounders, we show that our methods achieved lower estimation errors in counterfactual covariates and the most effective treatment timing than the baselines. Furthermore, using real basketball data, our methods performed realistic counterfactual predictions and evaluated the counterfactual passes in shot scenarios. 
\end{abstract}


%
\IEEEpeerreviewmaketitle

\section{Introduction}
\begin{figure}[t]
\centering
\includegraphics[width=1\linewidth]{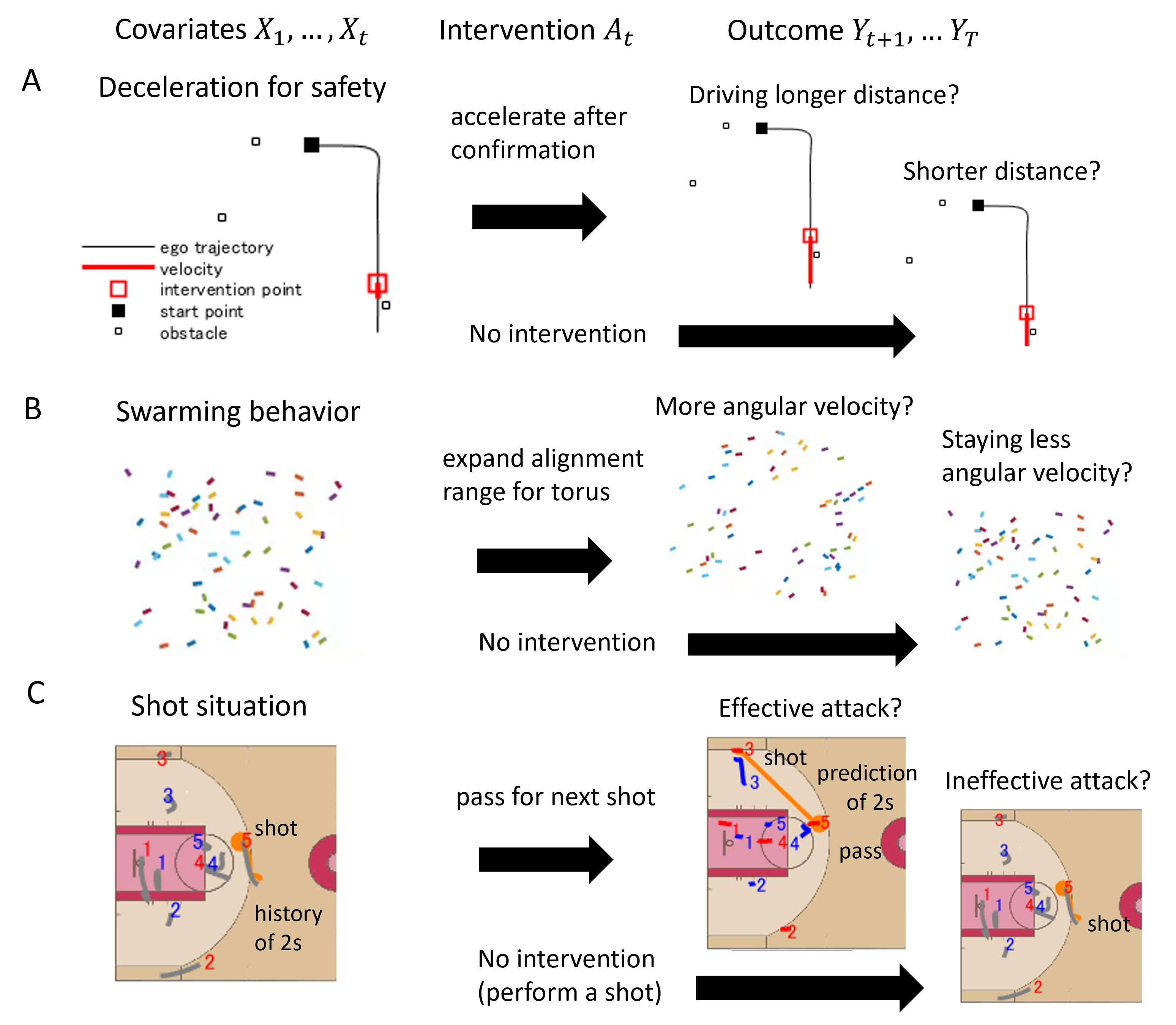}
\caption[]{The illustrations of our problems. Interventions in (A) an autonomous vehicle simulation, (B) a biological agent simulation, and (C) a real basketball are shown.
In (A) and (C), a single agent (A: an ego-vehicle and C: a ball player) is intervened whereas multiple agents (all boids) are intervened in (B).
\textcolor{black}{The motivations and variable definitions are described in the Introduction and Background sections.
In short, we aim to perform long-term counterfactual prediction of outcomes and covariates from the past covariates and intervention (or treatment assignment).}
In (A), our approach can test autonomous driving software including human interventions without creating the same situations for controlled trials (as real-world scenarios).
\textcolor{black}{The outcome is a safe driving distance.}
In (B), our approach has the potential to estimate the effect of an experimenter’s interventions on multi-animal behaviors, which improves the efficiency of experimental procedures for observing desired movements.
\textcolor{black}{The outcome is the angular velocity of multiple agents.}
In (C), our approach can estimate the effect of the selection of passes in basketball shot scenarios, thus we can evaluate the decision-making skills in this situation (e.g., during a game). \textcolor{black}{The outcome is the effectiveness of an attack.}

}

\label{fig:intervention}
\end{figure}

%
%
%
%
Evaluation of intervention in a real-world multiagent system is a fundamental problem in a variety of engineering and scientific fields. 
For example, a human driver in an autonomous vehicle, a player in team sports, and an experimenter on animals intervene in multiagent systems to obtain desirable results (e.g., safe driving, a good shot, and specific behavior, respectively) as shown in Fig. \ref{fig:intervention}.  
In these processes and complex interactions between agents, it is often difficult to estimate the intervention (or treatment) effect to compare the outcomes with and without interventions. 
There have been many methods to estimate the individual treatment effect (ITE), which evaluates the causal effect of treatment strategies on some important outcomes at the individual level in various fields (e.g., \cite{glass2013causal,baum2015causal,wang2015robust}).
In particular, some work has been proposed for dealing with time-varying \cite{lim2018forecasting, bica2020estimating} and hidden confounders \cite{bica2020time,liu2020estimating,ma2021deconfounding}.

However, most of the conventional frameworks did not consider the time-varying complex structures of multiagent relationships and counterfactual covariate predictions, which may lead to erroneous assessments of ITE and difficulty of interpretation. 
The structures of multiagent relationships include bottom-up ones based on interactions between local agents (often represented as a graph), and top-down ones described by global statistics and/or theories with domain knowledge \cite{Vicsek95,Couzin02}.
Since real-world multiagent systems do not usually have explicit governing equations, both top-down and bottom-up approaches can be simultaneously required for the modeling \cite{Fujii18,Fujii20}.
That is, in our problem, we need to model multiagent systems in both data-driven and theory-based approaches. 
In addition, for extracting insights from the ITE estimation results, the circumstances where intervention is effective must be analyzed.   
Therefore, plausible and interpretable modeling of time-varying multiagent systems is required for estimating ITE.  

In this paper, we propose a novel causal inference framework called Theory-based and Graph Variational Counterfactual Recurrent Network (\modelnospace), which estimates the effect of intervention in multiagent systems in interpretable ways.
The causal graph of the problem setting is illustrated in Fig. \ref{fig:causalgraph}, where the hidden confounders at a particular time stamp not only have causal relations to the observed variables at the same time stamp but also are causally affected by the hidden confounders from previous time stamps.
To model the hidden confounders and other causal relationships from data for the ITE estimation framework, we leverage graph variational recurrent neural networks (GVRNNs) \cite{Yeh19} to represent local agent interactions and theory-based computation for global properties of the multiagent behaviors based on domain knowledge.
This framework is based on the long-term prediction of multiagent covariates and outcomes can confirm under what circumstances the intervention is effective.   

Our general motivation is to estimate ITE over time in complex multiagent scenarios.
Specifically, in team sports, decision-making skills (e.g., whether a player with the ball should perform a pass or shot) are important. However, we cannot observe both patterns as data (i.e., performing a pass \textit{and} a shot) in the same situation. 
For autonomous driving or animal behavioral science, in real-world scenarios, we cannot obtain both data (i.e., with and without intervention) in the same situation (note that in the numerical experiment, we used synthetic data including both cases). For animal behavioral science, similarly, scientific experiments can hardly be performed with all possible intervention timing during movements and sometimes include the bias of the experimenters. 

In summary, our main contributions are as follows.
(1) We proposed a novel counterfactual recurrent network called \modelnospace, which estimates the effect of intervention in \textcolor{black}{\textit{multiagent systems in interpretable ways}, compared with the previous counterfactual recurrent networks dealing with time-varying treatment \cite{lim2018forecasting, bica2020estimating,bica2020time,liu2020estimating,ma2021deconfounding}.} 
(2) Methodologically, for the ITE estimation framework based on long-term prediction of multiagent covariates and outcomes, our model leverages GVRNN to represent local agent interactions and theory-based computation, \textcolor{black}{which was not considered in the above previous work. This framework} can confirm under what circumstances the intervention is effective.
(3) In experiments using two simulated models of an automated vehicle and biological agents, we show that our methods achieved lower errors in estimating counterfactual covariates and the most effective treatment timing than the baselines. Furthermore, using real basketball data, our methods performed realistic counterfactual predictions.
\textcolor{black}{All of these subjects moved interactively in multiagent systems, which are not dealt with in the previous work.}
We extend our previous short paper \cite{fujii2022estimating} by adding theoretical background, experimental results of a synthetic Boid dataset and a real-world basketball dataset, and a sensitivity analysis using the CARLA dataset.   

The remainder of this paper is organized as follows. First, in Section \ref{sec:problem}, we describe our problem definition. Next, we describe our methods in Section \ref{sec:proposed}.
We overview the related works in Section \ref{sec:related}, present experimental results in Section \ref{sec:experiments}, and conclude this paper in Section \ref{sec:conclusion}.

\section{Background} 
\label{sec:problem}
In this section, we first give definitions of the notations used throughout the paper, present the assumptions of our methods for estimating ITE, \textcolor{black}{and then introduce VRNNs (variational recurrent neural networks) \cite{Chung15} and GNN (graph neural network) \cite{Kipf18} we used}.

\subsection{Preliminary}
The multiagent observational data is denoted as the following $X_t, A_t, Y_t$ at time stamp $t$. 
Examples of these variables are shown in Fig. \ref{fig:intervention}\textcolor{black}{.}
Let $X_t$ be the time-dependent covariates of the observational data such that $X_t=\{x^{(1)}_{t},...,x^{(n)}_{t}\}$, where the $x^{(i)}_{t}$ denotes the covariates for $i$-th multiagent sample with $K$ agents, and $n$ denotes the number of samples. 
The relationships between agents are represented by the theory-based computation and GVRNN described in Section \ref{sec:proposed}. 
Although some related papers \cite{bica2020estimating,liu2020estimating} consider the static covariates $C$, 
which do not change over time, here we do not explicitly consider $C$ because we can easily formulate and implement our methods to add $C$ by conditioning. 
The treatment (or intervention) assignments are denoted as $A_{t}=\{a^{(1)}_{t},a^{(2)}_{t},...,a^{(n)}_{t}\}$, where $a^{(i)}_{t}$ denotes the treatments assigned in the $i$-th sample. 
We consider $a^{i}_{t}\in\{0, 1\}$, where $1$ is considered treated whereas $0$ is the control (i.e., a binary treatment setting), and we estimate the effect of the treatment assigned at time stamp $t$ on the outcomes $Y_{t+1}=\{y^{(1)}_{t+1},y^{(2)}_{t+1},...,y^{(n)}_{t+1}\}$ at time stamp $t+1$.
Note that in observational data, a multiagent sample can only belong to one group (i.e., either a treated or control group), thus the outcome from the other group is always missing and referred to as counterfactual.
To represent the historical sequential data before time stamp $t$, we use the notation $\overline{X}_{t}=\{X_{1},X_{2},...,X_{t-1}\}$ to denote the history of covariates observed before time stamp $t$, and $\overline{A}_{t}$ refers to the history of treatment assignments. Combining all covariates and treatments, we define $\mathcal{H}^{(i)}_{t}=\{\overline{x}^{(i)}_{t},\overline{a}^{(i)}_{t}\}$ as all the historical data collected before time stamp $t$. 

We adopt the potential outcomes framework (e.g., \cite{rubin1978bayesian}) and extended by \cite{robins2009estimation} to account for time-varying treatments.
The potential outcome $y^{(i)}_{a_t=a,t+1}$ of the $i$-th sample given the historical treatment can be formulated as $y^{(i)}_{a_t=a,t+1}=\mathbb{E}[y|x^{(i)}_{t},\mathcal{H}^{(i)}_t,a_t=a]$, where $a=\{0,1\}$. 
Then the ITE on the temporal observational data is defined as:
\begin{small}
\begin{equation}
    \tau^{(i)}_t=\mathbb{E}[y^{(i)}_{a_t =1,t+1}-y^{(i)}_{a_t =0,t+1}|x^{(i)}_{t},\mathcal{H}^{(i)}_{t}].
\end{equation}
\end{small}
Here, the observed outcome $y^{(i)}_{a_t=a,t+1}$ under treatment $a$ is called factual outcome, while the unobserved one $y^{(i)}_{a_t=1-a,t+1}$ is the counterfactual outcome.

\begin{figure}[t]
\centering
\includegraphics[width=0.8\linewidth]{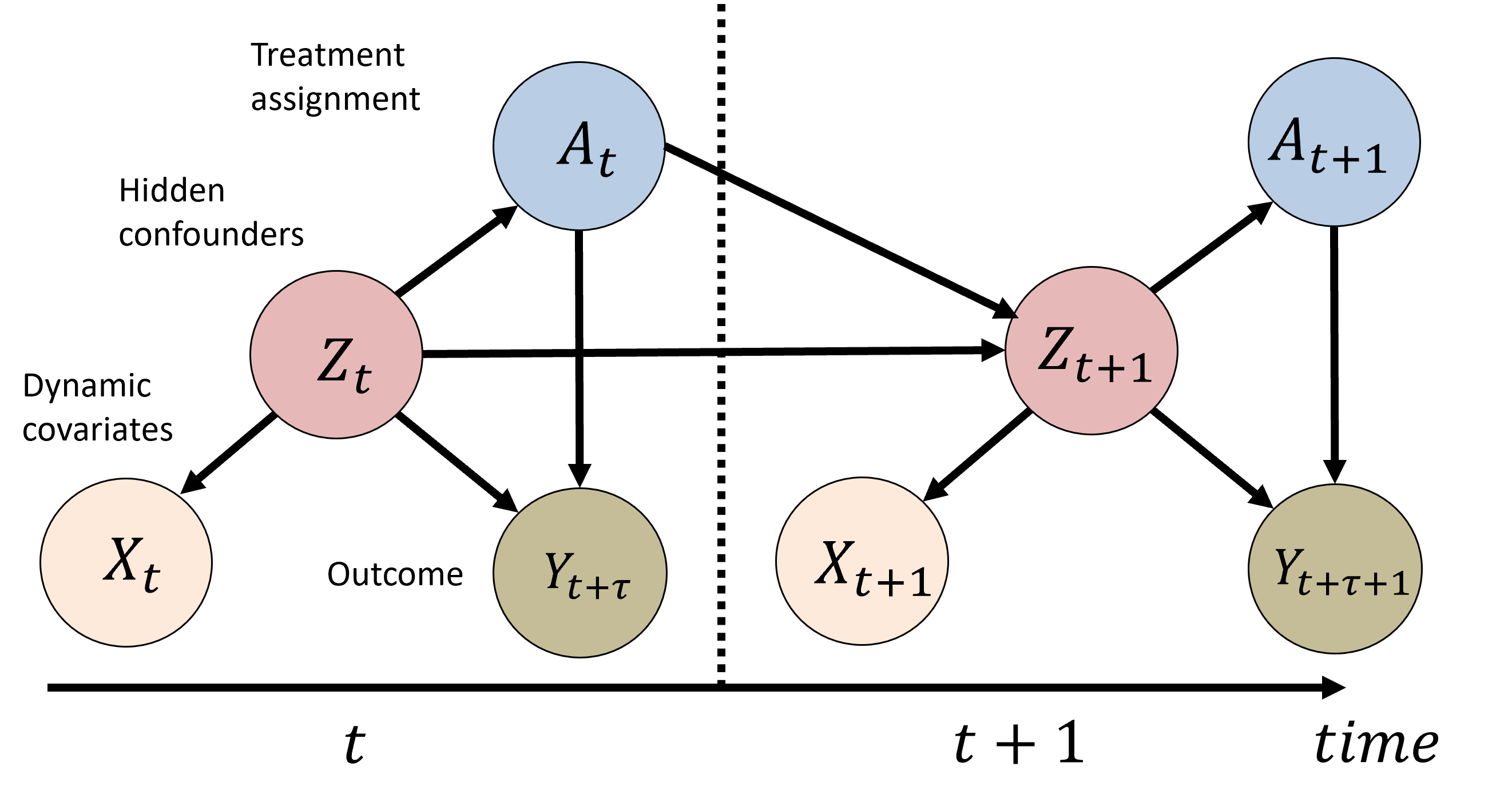}
\caption[]{The illustration of causal graphs for our problem.
We denote $X_t, Z_t,  A_t, Y_{t+1}$ as the dynamic covariates, representations of hidden confounders, treatment assignment, and outcomes, respectively. 
The black lines indicate the causal relations.
The hidden confounders $Z_{t+1}$ usually affect the treatment assignment $A_{t+1}$, the outcome $Y_{t+2}$, and the covariate $X_{t}$. To infer $Z_{t+1}$, we can leverage the observational data $X_{t+1}$ and previous hidden counfounders $Z_{t}$.
}

\label{fig:causalgraph}
\end{figure}

\subsection{Assumptions}
Our estimation of ITE is based on the following standard assumptions \cite{robins2000marginal, lim2018forecasting,hernan2010causal}, and we further extend the assumptions in our scenario to include time-varying observational data.

\begin{assumption}[Consistency]\label{as:consistency}
The potential outcome under treatment history $\overline{A}$ is equal to the observed outcome if the actual treatment history is $\overline{A}$.
\end{assumption}

\begin{assumption}[Positivity]\label{as:positivity}
For any sample $i$, if the probability  $p(\overline{a}^{(i)}_{t-1},\overline{x}^{(i)}_{t})\neq0$, then the probability of receiving treatment $0$ or $1$ is positive, i.e., 
$0<p(\overline{a}^{(i)}_{t},\overline{x}^{(i)}_{t} )<1$, for all $\overline{a}^{(i)}_{t}$. 
\end{assumption}
Assumption \ref{as:positivity} means that, for each time $t$, each treatment has a non-zero probability of being assigned. 
Besides these two assumptions, much of the existing work is based on the \textit{strong ignorability} assumption as follows:

\begin{definition}[Sequential Strong Ignorability]\label{as:ignorability}
Given the observed historical covariates $\overline{x}^{(i)}_t$ 
of the $i$-th sample, the potential outcome variables $\{y^{(i)}_{a_t=0,t+1}, y^{(i)}_{a_t=1,t+1}\}$ are independent of the treatment assignment, i.e., $\{y^{(i)}_{a_t=0,t+1}, y^{(i)}_{a_t=1,t+1}\}\indep  a^{(i)}_{t}|\overline{x}^{(i)}_t$. 
\end{definition}

Definition \ref{as:ignorability} means that there are no hidden confounders, i.e., all covariates affecting both the treatment assignment and the outcomes are present in the observational dataset. 
However, this condition is difficult to be guaranteed in practice especially in real-world observational data (in other words, it is not testable in practice \cite{robins2000marginal,  pearl2009causality}).
In this paper, we relax such a strict assumption by acknowledging potential hidden confounders. Our proposed methods can learn the representations of the hidden confounders and eliminate the bias between the treatment assignments and outcomes at each time stamp.

In our approach, the learned representations (denoted by $Z_{t}=\{z^{(1)}_{t},z^{(2)}_{t},...,z^{(n)}_{t}\}$) can be leveraged to infer the unobserved confounders and act as substitutes of hidden confounders. 
That is, we extend the strong ignorability assumption by considering the existence of hidden confounders $Z_{t}$, which influence the treatment assignment $A_t$ and potential outcomes $Y_{t+1}$. 
Given the hidden confounders $Z_{t}$, the potential outcome variables are independent of the treatment assignment at each time stamp. 
We aim to learn the representations of hidden confounders $Z_{t}$ for bias elimination based on the following assumptions \cite{ma2021deconfounding}: 

\begin{assumption}[Existence of Hidden Confounders]\label{as:hiddenconfounders}
(i) The hidden confounders may not be accessible, but the covariates are correlated with the hidden confounders, and can be considered as proxy variables, and
(ii) hidden confounders at each time stamp are also influenced by the hidden confounders and treatment assignments from previous time stamps.
\end{assumption}

Based on the premise, we study the identification of ITE: 
\setcounter{theorem}{0}
\begin{theorem}[Identification of ITE]\label{thm:identification}
If we recover $p(z_t^{(i)}|x_t^{(i)},\mathcal{H}_t^{(i)})$ and $p(y_{t+1}^{(i)}|z_t^{(i)},a_t^{(i)})$, then the proposed methods can recover the ITE under the causal graph in Fig. \ref{fig:causalgraph}.
\end{theorem}
\noindent We provide a proof in Appendix \ref{app:proof}.
For simplicity, the sample superscript ($i$) will be omitted unless explicitly needed.

\begin{figure*}[t]
\centering
\includegraphics[width=1\textwidth]{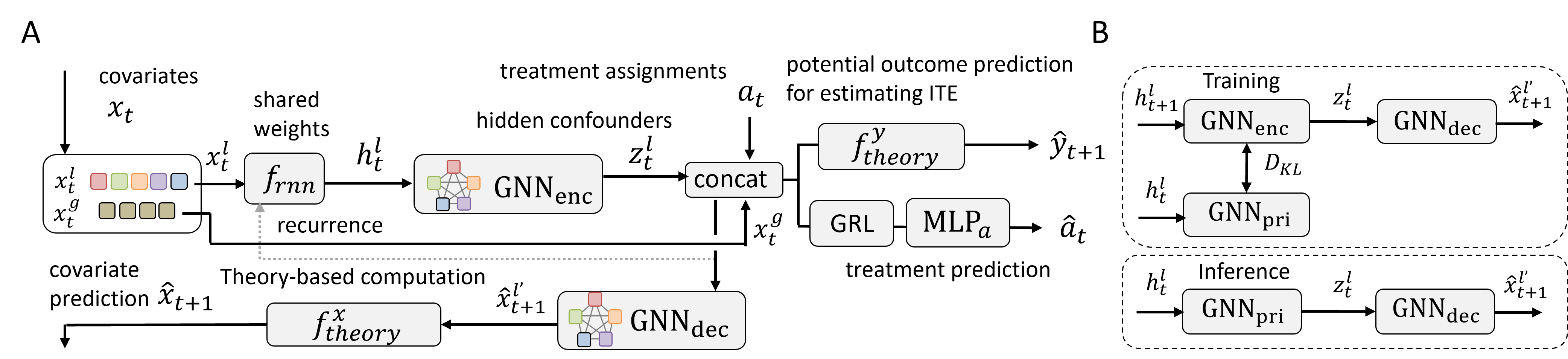}
\caption[]{The illustration of \model. 
\textcolor{black}{(A) \model aims to estimate ITE based on long-term prediction of multiagent covariates and outcomes while visualizing the long-term future covariate prediction. \model leverages GVRNN to represent local agent interactions and theory-based functions for covariate and outcome prediction, which can confirm under what circumstances the intervention is effective. Specifically, (B) the training and inference processes of GNN encoder, prior, and decoder are illustrated.}
At each time stamp, the model takes the current covariates and treatment assignments as input to learn representations of the hidden confounders via GRNNs and GNN encoders. 
Then, via theory-based computations, the GNN decoders, and MLPs (multi-layer perceptron), the model predicts time-varying covariates, a potential outcome, and a treatment. \textcolor{black}{We also use the gradient reversal layer before the treatment classifier to ensure the confounder representation distribution of the treated and that of the controlled are similar at the group level.}
}

\label{fig:model}
\end{figure*}

\textcolor{black}{
\subsection{Variational recurrent and graph neural networks}
}
\textcolor{black}{Here we explain VRNN \cite{Chung15} and GNN \cite{Kipf18} used in the following section \ref{sec:proposed}. }

\textcolor{black}{
\noindent{\bf{VRNN.}}
Let $x_{\leq T} = \{ x_1, \dots, x_T \}$ denote a sequence of variables of length $T$. 
The goal of sequential generative modeling is to learn the distribution over sequential data $\mathcal{D}$ consisting of multiple demonstrations. 
A common approach to model the trajectory is to factorize the joint distribution and then maximize the log-likelihood 
%
$\theta^* = \argmax_{\theta} \sum_{x_{\leq T} \in \mathcal D} \sum_{t=1}^T \log p_{\theta} (x_t | x_{<t}),$ 
%
where $\theta$ denotes the learnable parameters of models such as RNNs.
However, RNNs with simple output distributions often struggle to capture highly variable and structured sequential data (e.g., multimodal behaviors) \cite{Zhan19}. 
Recent work in sequential generative models addressed this issue by injecting stochastic latent variables into the model and optimization using amortized variational inference to learn the latent variables (e.g., \cite{Chung15,Fraccaro16,Goyal17}). %
Among these methods, a variational RNN (VRNN) \cite{Chung15} has been widely used in base models for multiagent trajectories \cite{Yeh19,Zhan19} with unknown governing equations.
A VRNN is essentially a variational autoencoder (VAE) conditioned on the hidden state of an RNN and is trained by maximizing the (sequential) evidence lower-bound (ELBO):
}
\textcolor{black}{
\eq{
\mathcal{L}_{vrnn} =  &  \mathbb{E}_{q_{\phi}(z_{\leq T} \mid x_{\leq T})}\Bigg[ \sum_{t=1}^T \log p_{\theta}(x_t \mid z_{\leq T}, x_{<t}) \nonumber \\
& - D_{KL} \Big( q_{\phi}(z_t \mid x_{\leq T}, z_{<t}) || p_{\theta}(z_t \mid x_{<t}, z_{<t}) \Big) \Bigg], \label{eq:vrnn_elbo} 
}}
\textcolor{black}{where $z_t$ is a stochastic latent variable of VAE, and
$p_{\theta}(x_t \mid z_{\leq t}, x_{<t})$, $q_{\phi}(z_t \mid x_{\leq t}, z_{<t})$, and $p_{\theta}(z_t \mid x_{<t}, z_{<t})$ are generative model, the approximate posterior or inference model, and the prior model, respectively.
The first term is the reconstruction term.
The second term is the Kullback-Leibler (KL) divergence between the approximate posterior and the prior.}

\textcolor{black}{
\noindent{\bf{GNN.}}
We then overview a graph neural network (GNN) based on \cite{Kipf18}.
Let $v_k$ be a feature vector for each node $k$ of $K$ agents. 
Next, a feature vector for each edge $e_{(k,j)}$ is computed based on the nodes to which it is connected. 
The edge feature vectors are sent as ``messages'' to each of the connected nodes to compute their new output state $o_k$.
Formally, a single round of message passing operations of a graph net is characterized below:
}
\textcolor{black}{
\eq{
v\rightarrow e: e_{(k,j)} &= ~f_e([v_k,v_j]), \\
e\rightarrow v: ~~~~~~o_i &= ~f_v\left(\sum_{j\in N(k)} e_{(k,j)}\right),
\label{eq:message} 
}}
\textcolor{black}{where $N(k)$ is the set of neighbors of node $k$ and $f_e$ and $f_v$ are neural networks.
In summary, a GNN takes in feature vectors $v_{1:K}$ and outputs a vector for each node $o_{1:K}$, i.e., $o_{1:K} = {\rm{GNN}}(v_{1:K})$. 
The operations of the GNN satisfy the permutation equivariance property as the edge construction is symmetric between pairs of nodes and the summation operator ignores the ordering of the edges \cite{zaheer2017deep}.
}

\vspace{10pt}
\section{Proposed Method}
\label{sec:proposed}
Here, we describe our \model method for ITE estimation in multiagent observational data. 
The overall framework is illustrated in Fig. \ref{fig:model}A.
We aim to combine predictions of outcome and covariates using data-driven and theory-based approaches, while balancing the representations of treated and control groups to reduce the confounding bias.
To this end, we first introduce the representation learning of hidden confounders with balancing by mapping the current multiagent
observational data and historical information into the representation space.
Next, we describe the prediction methods of the time-varying covariates, a potential outcome, and the treatment using the learned representations.
Finally, we describe the loss function.

\subsection{Representation learning of confounders}
\label{ssec:representation}
\textcolor{black}{Here, as a main approach in Fig. \ref{fig:model}A, we extend} a GVRNN \cite{Yeh19} for local multiagent locations $x^l_t$  (i.e., specific for each agent) with theory-based computation.
\textcolor{black}{As its variant (e.g., used for the ablation study)}, a pure data-driven model combining GVRNN and VRNN \cite{Chung15} for global variables $x^g_t$ (i.e., common for all agents) is also considered. 
\textcolor{black}{Since the global variables do not usually have the graph structure, VRNN without GNN is suitable}. 
Here we describe the representation learning of hidden confounders.

\noindent{\bf{GVRNN.}}
We first describe Graph Variational RNN (GVRNN) \cite{Yeh19} to obtain the representation from multiagent locations, which models the interactions between them at each step using GNNs. 
Let $x^l_{\leq T} = \{ x^l_1, \dots, x^l_T \}$ denote a sequence of covariates (here we consider multiagent locations). 
In this paper, GVRNN's update equations are as follows:
\eq{
p_{\theta}(z^l_t | x^l_{\leq t}, z^l_{<t}) & = \prod_k \mathcal{N}(z^l_{t,k}|\mu^{\rm{pri}}_{t,k},(\sigma^{\rm{pri}}_{t,k})^2), \\
q_{\phi}(z^l_t | x^l_{\leq t+1}, z^l_{<t}) & = \prod_k \mathcal{N}(z^l_{t,k}|\mu^{\rm{enc}}_{t,k},(\sigma^{\rm{enc}}_{t,k})^2),\\
p_{\theta}(x^l_{t+1} | z^l_{\leq t}, x^l_{\leq t}) & = \prod_k \mathcal{N}(z^l_{t,k}|\mu^{\rm{dec}}_{t,k},(\sigma^{\rm{dec}}_{t,k})^2),\\
h^l_{t+1,k} & = f^l_{rnn}(x^l_{t+1,k}, z^l_{t,k}, h^l_{t,k}),
\label{eq:gvrnn_state}
}
where $h^l_t$ and $z^l_t$ are deterministic and stochastic latent variables, $\mathcal{N}(\cdot|\mu,\sigma^2)$ denotes a multivariate normal distribution with mean $\mu$ and covariance matrix diag($\sigma^2$), and
\eq{
[\mu^{\rm{pri}}_{t,1:K},\sigma^{\rm{pri}}_{t,1:K}] & = {\rm{GNN_{pri}}}(h^l_{t,1:K}),\\
[\mu^{\rm{enc}}_{t,1:K},\sigma^{\rm{enc}}_{t,1:K}] & = {\rm{GNN_{enc}}}([x^l_{t+1,1:K},h^l_{t,1:K}]), \\
[\mu^{\rm{dec}}_{t+1,1:K},\sigma^{\rm{dec}}_{t+1,1:K}] & = {\rm{GNN_{dec}}}([z^l_{t,1:K},h^l_{t,1:K}]).
}
The prior network ${\rm{GNN_{pri}}}$, encoder ${\rm{GNN_{enc}}}$, and decoder ${\rm{GNN_{dec}}}$ are GNNs with learnable parameters $\phi$ and $\theta$.
\textcolor{black}{The relationship among them is illustrated in Fig. \ref{fig:model}B.}
Here we used the mean value $\mu^{\rm{dec}}_{t+1,1:K}$ as input variables $\hat{x}^{l'}_{t+1}$ in the following theory-based computation.
GVRNN is trained by maximizing the sequential ELBO in a similar way to VRNN as described in Eq. (\ref{eq:vrnn_elbo}), which is denoted as $\mathcal{L}_{gvrnn}$.

\noindent{\bf{Combined representation learning.}}
To construct a fully data-driven model (instead of the theory-based computation in Fig. \ref{fig:model}A), we propose a hierarchical GVRNN combining GVRNN for multiagent locations and VRNN for global inputs to learn the representation of hidden confounders.
In summary, each agent’s trajectory and other global information are processed through gated recurrent units (GRUs). 
The GRU parameters for the agent's trajectory are shared but keep its own individual recurrent state. 
At each time stamp, the model takes the current covariates and treatment assignments as input for learning representations of hidden confounders $z_t = [z^l_t,z^g_t]$ via GRUs, GNN, and MLP encoders. 

\subsection{Prediction with learned representation}
Our methods predict time-varying covariates, potential outcomes, and treatment for balancing by combining data-driven and theory-based approaches.
Here our contributions are to propose the theory-based computation of global variables and the prediction of time-varying covariates.
In this subsection, we describe the theory-based computation and the prediction of time-varying covariates, potential outcomes, and treatment for balancing.

\noindent{\bf{Theory-based computation.}} 
Here we assume that we do not have simulators including governing equations (used e.g., in  \cite{Kipf18,takeishi2021physics}) of multiagent systems such as team sports.  
In such a situation, we can utilize theory or prior knowledge of the domain using two approaches. 
One is to partially incorporate rule-based models into the data-driven model such as a mathematical relationship \cite{Fujii20policy} (e.g., between position and velocity), the biological constraints (e.g., turn angle in Sec. \ref{ssec:synthetic}), and critical behaviors (e.g., collision in Sec. \ref{ssec:synthetic}, ball movement in the air, and defending against the shot in Sec. \ref{ssec:basketball}).
Another is to compute auxiliary features for potential outcome predictions such as global variables (e.g., mean angular momentum in Sec. \ref{ssec:synthetic} or specific inter-agent distances in Secs. \ref{ssec:synthetic} and \ref{ssec:basketball}).
In our problem, we need to predict potential outcomes and covariates using predicted local variables from the data-driven model in Sec. \ref{ssec:representation} as shown in Fig. \ref{fig:model}A.
That is, 
\eq{\hat{y}_{t+1} = f^y_{theory}([z^l_t,x^g_t,a_t])\label{eq:theoryy}\\
\hat{x}_{t+1} = f^x_{theory}(\hat{x}^{l'}_{t+1}),\label{eq:theoryx}
}
where $\hat{x}_{t+1} = [\hat{x}^l_{t+1},\hat{x}^g_{t+1}]$ and $f^y_{theory}$ and $f^x_{theory}$ are theory-based functions utilizing domain knowledge. 
For the details, see Sec. \ref{sec:experiments}.

\noindent{\bf{Prediction of time-varying covariates.}} 
Another contribution of this paper is to propose methods to predict time-varying covariates, which can confirm under what circumstances the intervention is effective.
The proposed \model infers time-varying covariates via GNN decoders and the theory-based computation as illustrated in Fig. \ref{fig:model}A.
For a fully data-driven model, we use the VRNN decoder for global variables instead of the theory-based computation.
We minimize the factual covariate loss function as follows:
\begin{equation}
\label{eq:covariate_loss}
    \mathcal{L}_x=\frac{1}{nT}\sum^{n}\sum_{t=1}^{T}(\hat{x}_{t}-x_{t})^{2}.
\end{equation}

\noindent{\bf{Prediction of a potential outcome.}} 
Next, we describe a potential outcome prediction network to estimate the outcome $\hat{y}_{t+1}$ as described in Eq. (\ref{eq:theoryy}).
Depending on the problem of various data domains, MLP was placed before or after the theory-based function $f^y_{theory}$ to properly model the potential outcome (for details, see Section \ref{sec:experiments}).
For a fully data-driven model, we use MLP instead of the theory-based computation.
We minimize the factual loss function as follows:
\begin{equation}
\label{eq:outcome_loss}
    \mathcal{L}_y=\frac{1}{nT}\sum^{n}\sum_{t=1}^{T}(\hat{y}_{t+1}-y_{t+1})^{2}.
\end{equation}

\noindent{\bf{Treatment prediction and balancing.}} 
We also predict the treatment assignments $\hat{a}_t$ at each time stamp. The predicted treatments $\hat{a}_t$ are obtained through a fully-connected layer with a sigmoid function as the last layer.
That is, $\hat{a}_t$ is the probability of receiving treatment based on the confounders at time $t$, which can be typically referred to as propensity score \cite{rosenbaum1983central} $\hat{a}_t=p(a_{t}=1|z_{t})$.

Since we consider the binary treatment in this paper, we use a cross-entropy loss for the treatment prediction as follows:
\begin{equation}
\label{eq:treatment-pred-loss}
    \mathcal{L}_a=-\frac{1}{nT}\sum^{n}\sum_{t=1}^{T}(a_t\log{\hat{a}_t}+(1-a_t)\log{(1-\hat{a}_t)}).
\end{equation}

\begin{figure}[h]
\centering
\includegraphics[width=0.5\textwidth]{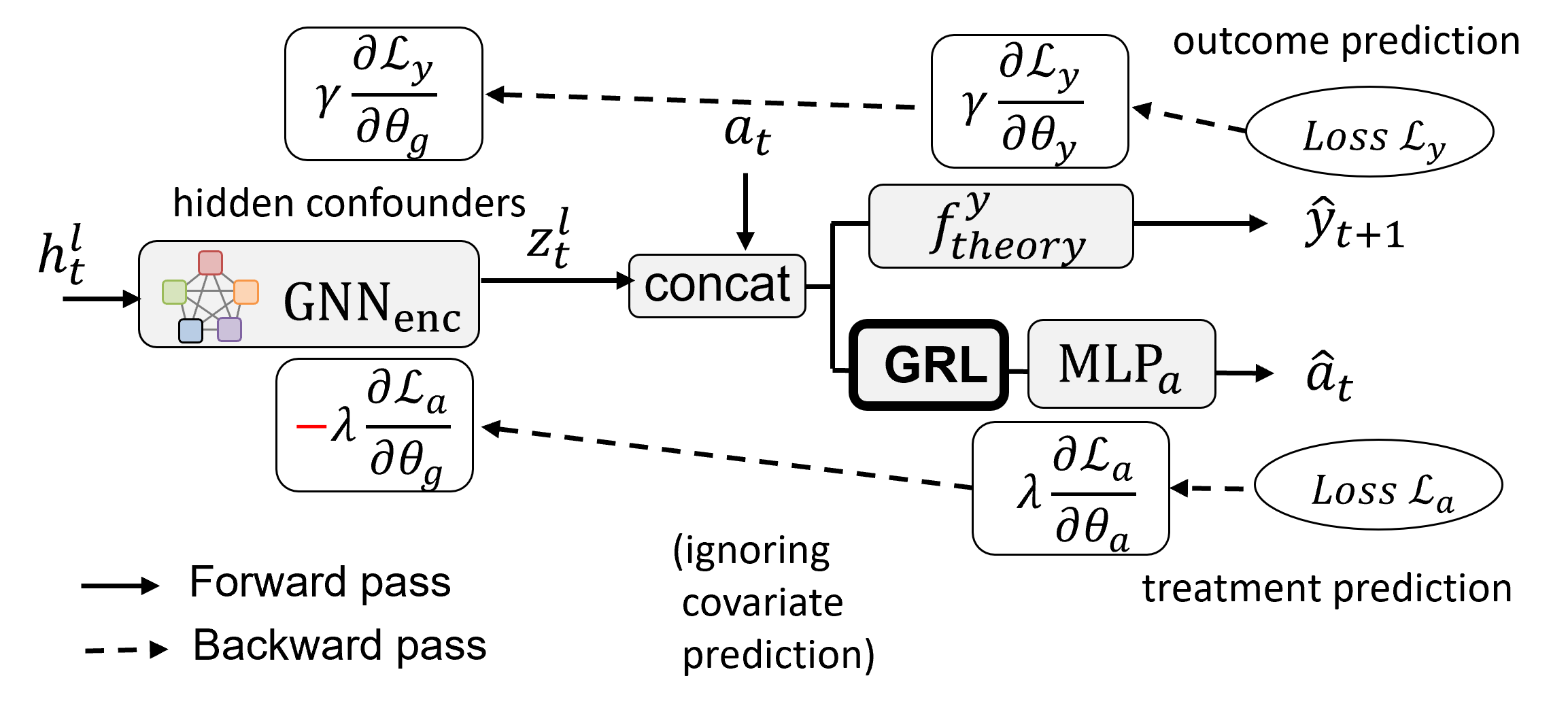}
\caption[]{\textcolor{black}{Illustration of gradient reversal layer (GRL). The diagram is a part of Fig. \ref{fig:model} ignoring covariate prediction and includes backward passes. $\theta_g$, $\theta_a$, and $\theta_y$ are the neural network parameters of the GNN encoder, MLP for treatment and outcome prediction (if $f^y_{theory}$ is a neural network). 
GRL multiplies the gradient by a certain negative constant during the backpropagation-based training. }
}
\label{fig:GRL}
\end{figure}

One critical consideration is balancing the representations of treated and control groups, which helps reduce the confounding bias \cite{bica2020estimating,liu2020estimating,ma2021deconfounding} and minimize the upper bound of the outcome inference error \cite{shalit2017estimating}. 
In this paper, we incorporate the
adversarial learning-based balancing method using a gradient reversal layer \cite{ganin2016domain} according to the related studies \cite{bica2020estimating,ma2021deconfounding}.
\textcolor{black}{Figure \ref{fig:GRL} illustrates our adversarial training process.
Let $\rm{GNN_{enc}(\cdot ; \theta_g)}$ be the GNN encoder with parameters $\theta_g$.
Also, let $\rm{MLP_a(\cdot ; \theta_a)}$ and $f^y_{theory}(\cdot ; \theta_y)$ be the MLP for treatment and outcome prediction with parameters $\theta_a$ and $\theta_y$ (if $f^y_{theory}$ is a neural network). $\gamma$ and $\lambda$ are the hyperparameters (or weights) of loss functions as described below. 
}
We add the gradient reversal layer before the treatment classifier \textcolor{black}{$\rm{MLP_a(\cdot; \theta_a)}$} to ensure the confounder representation distribution of the treated and that of the controlled are similar at the group level \cite{bica2020estimating,ma2021deconfounding}. 
The gradient reversal layer does not change the input during the forward-propagation, but in the back-propagation, reversing the gradient by multiplying by a negative scalar \textcolor{black}{(i.e., $-\lambda \frac{\partial \mathcal{L}_a}{\partial \theta_g}$ in Fig. \ref{fig:GRL}}). 
Although the model will minimize the loss of the treatment prediction, the adversarial learning process will contribute to the distribution balancing during the prediction of the potential outcome and the covariates. 

\subsection{Loss function}
The loss function for our method is defined as
\begin{equation}
    \mathcal{L} = \mathcal{L}_y + \alpha \mathcal{L}_{gvrnn} + \gamma \mathcal{L}_x + \lambda \mathcal{L}_a 
    \label{eq:loss}
\end{equation}
where $\mathcal{L}_y$ is the factual outcome prediction loss, $\mathcal{L}_{gvrnn}$ is the ELBO in GVRNN, $\mathcal{L}_x$ is the covariate prediction loss, and $\mathcal{L}_a$ is the treatment prediction loss. 
$\alpha$, $\gamma$, and $\lambda$ 
are hyperparameters to balance the loss function. 
In a pure data-driven model, we add $\mathcal{L}_{vrnn}$ to $\mathcal{L}$.
To prevent over-fitting, we select the best-performing model using the loss function on the validation set.
The sensitivity analysis in the hyperparameters is presented in Appendix \ref{app:sensitivity}.

\section{Related work}
\label{sec:related}
\noindent\textbf{Learning causal effects with time-varying data.}
Pioneering work to estimate the effects of time-varying exposures has developed such as 
g-computation, structural nested models, and marginal structural models (e.g., \cite{robins1986new, robins1994correcting, robins2000marginal, robins2009estimation}) in statistics and epidemiology domains. 
To handle complex time-dependencies, Bayesian non-parametric approaches based on Gaussian processes \cite{xu2016bayesian,schulam2017reliable,soleimani2017treatment} and Dirichlet processes \cite{roy2017bayesian} have been proposed. 
Moreover, to model time-dependencies without strong assumptions on functional forms, RNN approaches have been intensively investigated such as recurrent marginal structural networks \cite{lim2018forecasting} and adversarial training to balance the historical confounders \cite{bica2020estimating}. 
Recently, to estimate the treatment effects with hidden confounders, several methods have been proposed by relaxing the assumption of strong ignorability \cite{bica2020time,liu2020estimating,ma2021deconfounding}.
Our approach is related to these works, but they did not consider the utilization of domain knowledge and the explicit prediction of time-varying covariates, which is necessary for the proper interpretation of the results.

\noindent\textbf{Representation learning for treatment effect estimation.}
Estimation methods of balanced representation between treated and control groups in hidden space have been proposed in the static setting \cite{hill2011bayesian,wager2018estimation,alaa2018limits}.
In neural network approaches, methods with regularization for the balancing \cite{johansson2016learning,shalit2017estimating}, incorporating the local similarity among individuals \cite{yao2018representation},  generative adversarial nets (GAN) \cite{yoon2018ganite}, a multi-task learning \cite{shi2019adapting}, \textcolor{black}{probabilistic 
 modeling \cite{grecov2022probabilistic,abroshan2022conservative}, and optimal transport framework \cite{li2021causal}} have been proposed.  
In the dynamic setting, \cite{bica2020estimating,ma2021deconfounding} adopted adversarial training techniques with a gradient reversal layer, \cite{bica2020time} proposed multi-task learning to build a factor model of the cause distribution, and \cite{liu2020estimating} used inverse probability of treatment weighting. 
\textcolor{black}{To model multiagent or networked systems, GNN-based approaches have been intensively used in prediction problems (e.g., \cite{wu2019graph, zhang2020spatio}).} 
To obtain representations from the networked covariates, \cite{ma2021deconfounding} incorporated graph convolutional networks \textcolor{black}{and \cite{takeuchi2023causal} incorporated hierarchical spatial graph structure into causal estimation frameworks}
(other counterfactual learning on graphs was surveyed by \cite{guo2023counterfactual}). Recently, continuous-time causal inferences for dynamical systems \cite{seedat2022continuous,jiang2023cf} have been proposed.
We adopted the adversarial training to balance the representation and introduced GVRNN with prior (domain) knowledge for a more accurate long-term prediction of potential outcomes and covariates.


\noindent\textbf{Counterfactual prediction for multiagent data.}
In past research of real-world multiagent movements, counterfactual prediction of human movements has been studied.
\textcolor{black}{As control problems, forward models such as in robotics, can be considered. However, our problem is inherently a backward problem to model the multiagent behaviors from data, then here we focus on such backward approaches.}
Counterfactual trajectory prediction methods have been proposed in pedestrians \cite{chen2021human}, animals \cite{fujii2021learning}, and team sports \cite{Yeh19,Fujii20policy,nakahara2022estimating,teranishi2022evaluation}.
Outside the context of causal inference based on the Rubin causal model, the causality of the physics-based system such as using neural rational inference \cite{Kipf18} and learning physics in a game engine with randomized stability \cite{lerer2016learning} have been considered (sometimes called causal reasoning).
\textcolor{black}{In the context of reinforcement learning, the counterfactual action prediction and evaluation can be considered in team sports \cite{Liu2018,Liu2020,rahimian2022beyond,nakahara2023action}.}
In traffic environments, 
methods to understand the reasons behind the maneuvers of other vehicles and pedestrians for escaping accidents \cite{ramanishka2018toward, you2020traffic,mcduff2021causalcity} have been proposed.
In the static setting of causal inference, \cite{takeuchi2021grab} proposed a spatial convolutional counterfactual regression to estimate the effects of crowd movement guidance. 
In team sports, propensity score matching was used to investigate the causal effect of some plays or timeouts in many sports \cite{yam2019lost,toumi2019grapes,gibbs2020causal,nakahara2022pitching}.
In the dynamic setting, \cite{vock2018estimating} applied a g-computation method to examine the effect of a specific pitch in baseball. 
We firstly propose a framework for estimating ITE in multiagent motions. 

\section{Experiments}
\label{sec:experiments}
The purpose of our experiments is to validate the proposed methods for application to real-world multiagent trajectories, which usually have no ground truth of the interaction rules. 
Hence, for verification of our methods, we first compared their performances to infer the ITE to those in various baselines using two synthetic datasets with ground truth: the CARLA \cite{dosovitskiy2017carla} autonomous driving simulator and a biological multiagent model called Boid \cite{Couzin02}.
In particular, for real-world applications, we need long-term counterfactual predictions.
We mainly verified the estimation of ITE using the CARLA dataset (relatively fewer interactions) and that of the best intervention timing using the Boid dataset (relatively more and complex interactions) because the intervention timing is sensitive.
Finally, we examined the applicability of our methods to real-world data using the basketball (NBA) dataset.
In common, we separated the time $T$ into a burn-in period $1,\ldots,T_b$ and a prediction period $T_p = T_b+1,\ldots,T$.
The intervention point during the intervention period is denoted as $T_i$ (a similar interval to $T_p$, but we set it differently in each problem). 
The hyperparameters of the models were determined by validation datasets in each experiment (for the details, see Appendices \ref{app:boid}, \ref{app:carla}, and \ref{app:nba}). 

Here, we commonly compared our methods to four baseline methods: 
a simple RNN baseline using GRU \cite{Cho14},
deep sequential weighting (DSW) \cite{liu2020estimating} as a baseline considering hidden confounders, graph counterfactual recurrent network (GCRN: for clarity, we change the name) \cite{ma2021deconfounding} 
, and the variant to predict covariates (GCRN$+$X).
These baselines were modified to our setting (e.g., multiagent and time-varying outcome: for details, see Appendix \ref{app:common}). 
For verification, since the prediction of the covariates is required, the most appropriate baseline is GCRN$+$X, which is compared via visualization because DSW and GCRN cannot predict the covariates.  
We also validated our approach with three variants: GV-CRN (without the theory-based computation, i.e., without global covariates), TV-CRN (removing GNN), and TG-CRN (replacing VRNN with RNN).
\textcolor{black}{To perform fair comparisons among the models, we trained all models with 20 epochs (the learning curves are shown in Appendix \ref{app:converge}).}
Other common training details 
are described in Appendix \ref{app:common} and the codes are provided at \url{https://github.com/keisuke198619/TGV-CRN}.

\subsection{Synthetic datasets}
\label{ssec:synthetic}
\vspace{-0pt}
To verify our method, we compared the performances to infer the causal effect with those in various baselines using two synthetic datasets with ground truth.
\textcolor{black}{We used two types of simulations: autonomous vehicle and biological agent (Boid) simulations. All simulations are rule-based rather than learning-based as described below.}
As for performance metrics, we first adopted commonly used potential outcome and covariate prediction errors.
Potential outcome prediction errors were computed as an absolute error of all simulated outcomes: $L_{outcome}^{(a_t,t,i,t')} = |\hat{y}_{a_t,t}^{(i,t')} - y_{a_t,t}^{(i,t')} |$, where a treatment $a_t = \{0,1\}$, time $t = [T_b,T+1]$, simulated temporal variation of the treatment timing $t' \in T_i$, and the burn-in period in RNN $T_b$.
Covariate prediction errors were computed as $L_2$ prediction error of all simulated covariates: $L_{covariates}^{(a_t,t,i,t')} = \|\hat{x}_{a_t,t}^{(i,t')} - x_{a_t,t}^{(i,t')} \|_2$.
We took the average over all variables for evaluation.

We also adopted two widely-used evaluation metrics in causal inference: rooted precision in the estimation of heterogeneous effect (PEHE) \cite{hill2011bayesian} 
$\sqrt{\epsilon^t_{PEHE}}=\sqrt{\frac{1}{n}\sum_{i\in n}(\hat{\tau}_t^{(i)}-\tau_t^{(i)})^2}$
and mean absolute error of the average treatment effect (ATE) \cite{willmott2005advantages} to measure the quality of the estimated individual treatment effects at different time stamps: $\epsilon^t_{ATE}=|\frac{1}{n}\sum_{i\in n}\hat{\tau}_t^{(i)}-\frac{1}{n}\sum_{i\in n}\tau_t^{(i)}|$.
We took the average over all time stamps for evaluation.

\textcolor{black}{Regarding interpretability, we emphasize that there has been no previous work to visualize or interpret future covariates (i.e., multiagent trajectory), which can confirm under what circumstances the intervention is effective. 
We then illustrated trajectory prediction results in each domain.
We compared the customized baselines to predict future covariates to demonstrate the superiority of our approach, which are also quantitatively shown as the covariate losses $L_{covariates}$. }

\noindent {\bf{Autonomous vehicle simulation.}}
We first validated the performances of our methods using an autonomous vehicle simulation using the CARLA simulator \cite{dosovitskiy2017carla} (ver. 0.9.8). 
The CARLA environment contains dynamic obstacles (e.g., pedestrians and cars) that interact with the ego car. 
To generate data, we simulated autonomous driving at various towns, starting points, and obstacle positions, which were subsampled at 2 Hz (see also Appendix \ref{app:carla}).
Since we did not use the future path as inputs for generality, we randomly split 7 towns data into 904 training, 129 validation, and 259 test scenarios. 
Here we used two types of autonomous vehicles: partial observation with Autoware \cite{kato2015open} (ver. 1.12.0) and full observation types with a pre-installed CARLA simulator. 
The partial observation model often stops when dangerous situations arise for safety. 
The full observation model intervenes to accelerate the stopped vehicle after the safety confirmation. 
We set $T=60, T_b=40$, and $T_i=T_b+1,\ldots,T_b+10$. 
For counterfactual data, since dangerous situations (i.e., requiring intervention) are limited, we only used with or without intervention at the same timing (not various timings). 

We predict the safe driving distance of the ego car from the starting point as an outcome while driving without any collision with obstacles. 
That is, the treatment effect of the full observation model is defined as the difference between safe driving distances with and without interventions.
As the local covariates $x^l_t$, we used the position, velocity, pose, and size information of the ego car and obstacles in the 2D map.
The maximum number of dynamic obstacles was $120$, but since the obstacles related to the ego car were limited and the number changes over time, we used the nearest $10$ obstacles' information as the covariates. 
As the global covariates $x^g_t$, we used the current driving distance $x_t^{dist}$ and (binary) collision $x_t^{coll}$ information of the ego car.
The theory-based function $f^x_{theory}$ mathematically computed $\hat{x}_{t+1}$ using the predicted velocity and pose information, and added the learned positive velocity in intervention.
$f^y_{theory}$ computed $\hat{y}_{t+1}$ using the predicted $\hat{y}'_{t+1}$ as the output of the MLP such that $\hat{y}_{t+1} = (1-x_t^{coll})\hat{y}'_{t+1}$ based on the definition of the safe driving distance.

\begin{figure}[h]
\centering
\includegraphics[width=0.45\textwidth]{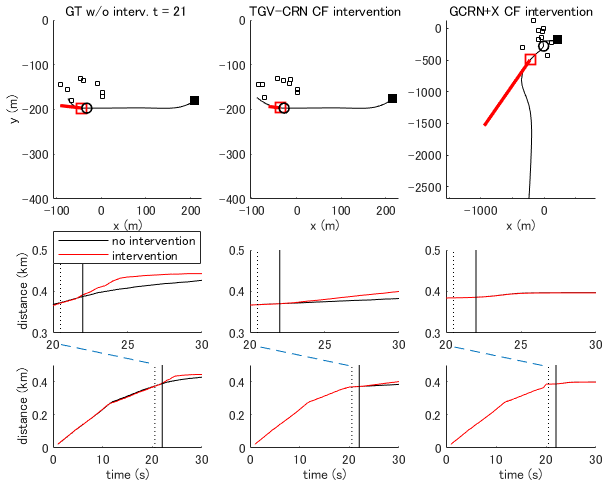}
\caption[]{Example CARLA results using our method. (Top) Visualization of covariates and (\textcolor{black}{middle row and} bottom) outcome time series in (left) ground truth without intervention, (middle \textcolor{black}{column}) counterfactual intervention using our model, and (right) the baseline. 
The middle \textcolor{black}{row} subfigures are enlarged views of the bottom ones \textcolor{black}{from 20 s}.
An ego car (red square) and obstacles (black) are shown in the upper plots (see also Fig. \ref{fig:intervention}A) at the intervention time, which is the solid line in the lower plots.
The unfilled circle is the start of the long-term prediction (dashed line in the lower plots).
The ego-car moves from right to left and stops because of the obstacles.
\textcolor{black}{The videos are given in the above GitHub page.}
}
\label{fig:carla}
\end{figure}

Quantitative verification results are shown in Table \ref{tab:carla}.
Our full model and its ablated variants show lower prediction errors of the outcome and covariates, and $\sqrt{\epsilon^t_{PEHE}}$, and $\epsilon^t_{ATE}$ than other baselines. 
Totally, our full model and the variant without amortized variational inference (TG-CRN) show the best performances, suggesting that the necessity of complex modeling for long-term prediction using this dataset would be smaller than that using the following datasets. 
In our four models, we found that on this dataset the combination of theory-based computation and GNN worked well in the covariate prediction, but did not in the outcome prediction.  
Example results of our method are shown in Fig. \ref{fig:carla}\textcolor{black}{, which can interpret the results}.
Our method (\model) shows better counterfactual prediction with intervention than the baseline (GCRN$+$X).
On average, the outcome values with intervention compared to the absence of the intervention (i.e., ITE) were $0.005 \pm 0.000$ (km) in our full model, $-0.021 \pm 0.000$ in the baseline (GCRN+X), and $0.013 \pm 0.001$ in the ground truth.
As Fig. \ref{fig:carla} shows, the baseline did not model the intervention effects.
\textcolor{black}{We expect that this method can be applied to the effect of human intervention in Level 3 autonomous vehicle simulations, which need human intervention. This approach can examine the effect of autonomous control in Level 4 or 5 autonomous vehicle simulations (without human intervention) in a case when human intervention is necessary in Level 3.}

\newcommand{\md}[2]{\multicolumn{#1}{c|}{#2}}
\newcommand{\me}[2]{\multicolumn{#1}{c}{#2}}
\begin{table}[ht!]
\centering
\scalebox{0.8}{
\begin{tabular}{l|cccc}
\Xhline{3\arrayrulewidth} 
& \me{1}{$L_{Outcome}$ } & \me{1}{$\sqrt{\epsilon^t_{PEHE}}$} &  \me{1}{$\epsilon^t_{ATE}$} & \me{1}{$L_{Covariates}$} \\
\hline
RNN & 3.330 $\pm$ 0.286 &  0.159 $\pm$ 0.025 & 0.129 $\pm$ 0.023 & 0.479 $\pm$ 0.0011
\\DSW \cite{liu2020estimating}& 0.161 $\pm$ 0.013 & 0.030 $\pm$ 0.003 & 0.019 $\pm$ 0.003  & ---
\\GCRN \cite{ma2021deconfounding} &  0.094 $\pm$ 0.009 & 0.024 $\pm$ 0.002 & 0.014 $\pm$ 0.002 & ---
\\GCRN$+$X &  3.020 $\pm$ 0.496 &  0.049 $\pm$ 0.009 & 0.034 $\pm$ 0.007 & 2.072 $\pm$ 0.0098\\
\hline
GV-CRN & \textbf{0.038} $\pm$ \textbf{0.003} &  0.022 $\pm$ 0.002 & 0.012 $\pm$ 0.001 & 0.372 $\pm$ 0.0010 
\\TG-CRN &  0.045 $\pm$ 0.004 &  \textbf{0.021} $\pm$ \textbf{0.002} & \textbf{0.011} $\pm$ \textbf{0.001} & \textbf{0.082} $\pm$ \textbf{0.0004}
\\TV-CRN & \textbf{0.032} $\pm$ \textbf{0.003} &  0.023 $\pm$ 0.002 & 0.014 $\pm$ 0.002 & 0.102 $\pm$ 0.0004 
\\ TGV-CRN (full) & 0.041 $\pm$ 0.003 &  \textbf{0.020} $\pm$ \textbf{0.002} & \textbf{0.008} $\pm$ \textbf{0.001} & \textbf{0.098} $\pm$ \textbf{0.0004}
\\
\Xhline{3\arrayrulewidth} 
\end{tabular}
}
\vspace{3pt}
\caption{\label{tab:carla} Performance comparison on the carla dataset. \textcolor{black}{The upper and lower rows indicate the baselines and proposed methods, respectively.}}
\end{table}

\begin{table*}[h!]
\centering
\scalebox{0.9}{
\begin{tabular}{l|ccccc|ccc}
\Xhline{3\arrayrulewidth} 
& \md{5}{Boid simulation dataset}& \me{3}{Real-world NBA dataset}  \\ 
&\me{1}{Treatment timing} & \me{1}{$L_{Outcome}$ } &
\me{1}{$\sqrt{\epsilon^t_{PEHE}}$} &  \me{1}{$\epsilon^t_{ATE}$}  & \md{1}{$L_{Covariates}$} &   \me{1}{$L_{Outcome}$ } & \me{1}{$\hat{\tau}^{CFP}$} & \me{1}{$L_{Covariates}$} \\ 
\hline
RNN & 1.815 $\pm$ 0.068 & 0.668 $\pm$ 0.033 &  0.443 $\pm$ 0.054 & 0.085 $\pm$ 0.016 & 0.168 $\pm$ 0.0002 & 0.291 $\pm$ 0.003 &  0.170 $\pm$ 0.0065&  0.969 $\pm$ 0.0024
\\DSW \cite{liu2020estimating}
& 2.353 $\pm$ 0.080 & \textbf{0.586} $\pm$ \textbf{0.029} & \textbf{0.440} $\pm$ \textbf{0.054} & \textbf{0.082} $\pm$ \textbf{0.015} & --- & 0.197 $\pm$ 0.003&  0.166 $\pm$ 0.0064 & ---
\\GCRN \cite{ma2021deconfounding} 
& 2.290 $\pm$ 0.081 & \textbf{0.587} $\pm$ \textbf{0.028} & 0.443 $\pm$ 0.055 & 0.088 $\pm$ 0.016 & --- & \textbf{0.170 }$\pm$ \textbf{0.003}&  0.149 $\pm$ 0.0062  & ---
\\GCRN$+$X & 2.290 $\pm$ 0.081 & 0.727 $\pm$ 0.022 &  \textbf{0.440} $\pm$ \textbf{0.054} & \textbf{0.080} $\pm$ \textbf{0.014} & 0.162 $\pm$ 0.0001 & 0.431 $\pm$ 0.004 &  0.154 $\pm$ 0.0063&  1.226 $\pm$ 0.0032\\
\hline
GV-CRN &  1.900 $\pm$ 0.068 & 0.674 $\pm$ 0.028 &  0.536 $\pm$ 0.045 & 0.090 $\pm$ 0.014 & 0.329 $\pm$ 0.0003 & 0.475 $\pm$ 0.005 &  0.151 $\pm$ 0.0060&  1.173 $\pm$ 0.0028
\\TG-CRN & \textbf{1.750} $\pm$ \textbf{0.062} & 0.694 $\pm$ 0.041 &  0.501 $\pm$ 0.067 & 0.125 $\pm$ 0.021 & \textbf{0.086} $\pm$ \textbf{0.0002} & 0.347 $\pm$ 0.005 &  0.205 $\pm$ 0.0072&  \textbf{0.811} $\pm$ \textbf{0.0022}
\\TV-CRN & \textbf{1.692} $\pm$ \textbf{0.064} & 0.981 $\pm$ 0.044 &  0.483 $\pm$ 0.052 & 0.106 $\pm$ 0.018 & 0.090 $\pm$ 0.0002  &  \textbf{0.165} $\pm$ \textbf{0.002} &  0.192 $\pm$ 0.0068&  0.852 $\pm$ 0.0023
\\ TGV-CRN (full) & 1.853 $\pm$ 0.062 & 0.690 $\pm$ 0.032 &  0.537 $\pm$ 0.055 & 0.101 $\pm$ 0.016 & \textbf{0.085} $\pm$ \textbf{0.0002} & 0.231 $\pm$ 0.004 &  0.261 $\pm$ 0.0083&  \textbf{0.825} $\pm$ \textbf{0.0023}
\\
\Xhline{3\arrayrulewidth} 
\end{tabular}
}
\vspace{3pt}
\caption{\label{tab:boid} Performance comparison on the boid and NBA datasets.\textcolor{black}{The upper and lower rows indicate the baselines and proposed methods, respectively.}}
\end{table*}

\noindent {\bf{Biological multiagent simulation.}}
Here, we validated our methods on the Boid model, which contains movement trajectories of 20 agents.
The Boid model (originally, \cite{reynolds1987flocks}) is a rule-based model to generate generic simulated flocking agents and we used a unit-vector-based (rule-based) model \cite{Couzin02} (for details, see Appendix \ref{app:boid}).
In this paper, we intervene the agents' recognition to generate torus (circle) behaviors from the swarm (random) behaviors.
The outcome is defined as the mean angular momentum of individuals about the center of the group (assuming the mass of each agent $m=1$). 
That is, the treatment effect of the change in the recognition is estimated as the difference in the future mean angular momentum between simulations with and without the interventions.

In this model, $20$ agents are described by a 2-D vector with a $1$ m/s constant velocity in a 15 $\times$ 15 m boundary square.
At each time stamp, a member will change direction according to the positions of all other members based on three zones.
The first is the repulsion zone with radius $r_r = 0.5$ m, in which individuals within each other’s repulsion zone try to avoid each other by swimming in opposite directions. 
The second is the orientation zone, in which individuals try to move in the same direction; 
here we set radius $r_o = 1$ to generate swarming behaviors before the intervention. 
To generate torus behaviors, we change $r_o = 4$, which is the intervention in this study.
The third is the attractive zone (radius $r_a = 7.5$ m), in which agents move towards each other and tend to cluster.

To simulate the treatment assignments, we generate factual $20800$ samples ($20000$ training, $400$ validation, and $400$ test datasets).
We set $T=14$ and $T_b = 9$ and randomly pick the intervention point during the intervention period $T_i = 9,\ldots,13$. 
The outcome is defined as the mean angular momentum among individuals at time $t+1$. 
We also created a counterfactual dataset only in the test dataset.
Here we created all combinations of treatment points during the intervention period $T_i$. 
As the local covariates $x^l_t$, we used position, velocity, and directional change of all agents.
As the global covariates $x^g_t$, we used the current mean angular momentum.
The theory-based function $f^x_{theory}$ mathematically computed $\hat{x}_{t+1}$ using the direction change $d$ at the next step, maximum turn angle $\beta$ as body constraints (for $d$ and $\beta$, see also Appendix \ref{app:boid}), and attraction rule when agents are far from the center of the group.
In addition, we added the orientation rule when agents are in the orientation zone and not in the repulsion zone in $f^x_{theory}$. 
$f^y_{theory}$ was replaced with a MLP such that $\hat{y}_{t+1} = {\rm{MLP_y}}([z^l_t,x^g_t,a_t])$.

\begin{figure}[h]
\centering
\includegraphics[width=0.45\textwidth]{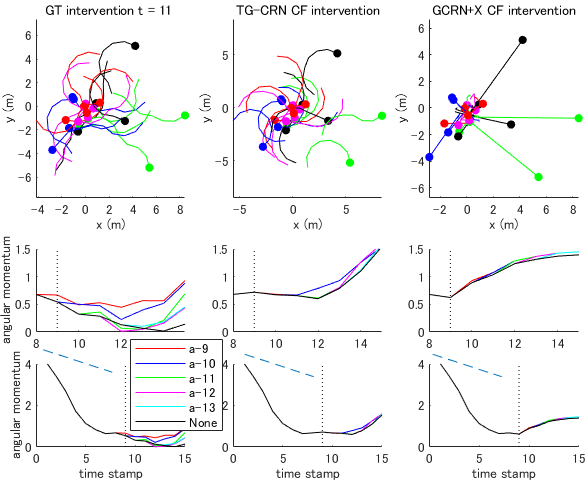}
\caption[]{Example Boid results of our method. The configurations are the same as Fig. \ref{fig:carla}. 
(Top) 20 boids with 6 different colors move in a rule-based manner (but we did not use this information for the prediction) and the filled circles are the starting point of the trajectories. 
(\textcolor{black}{Middle row and bottom) 
outcome time series in (left) ground truth without intervention, (middle column) counterfactual intervention using our model, and (right) the baseline. 
The middle row subfigures are enlarged views of the bottom ones from the 9th frame. }
The ``a'' in the lower caption is the intervention times.
\textcolor{black}{For example, ``$a = 9$'' means the case of intervention at the 9th frame and ``None'' indicates no intervention.
}
\textcolor{black}{The videos are given in the above GitHub page.}
}
\label{fig:boid}
\end{figure}

The results are shown in Table \ref{tab:boid} left. 
In addition to the four indices in the CARLA experiment, we investigated the estimation error of the best intervention timing $|\argmax_{t'}(\hat{y}^{(i,t')}_{T+1})-\argmax_{t'}(y^{(i,t')}_{T+1})|$.
The results indicate that our model and its variants with theory-based computation show better prediction performances in covariates and the best intervention timing than all of the baselines.
However, the outcome prediction errors in our models were worse than the causal inference baselines (DSW, GCRN, and GCFN+$X$), which may lead to degraded performances in  $\sqrt{\epsilon^t_{PEHE}}$ and $\epsilon^t_{ATE}$ of our models than the baselines. 
In our four models, all combination of core three components (T, G, and V) did not work well and on this dataset, TG-CRN without VRNN is the best performing model. 
On average, the ITE values were $0.160 \pm 0.010$ in our best model (TG-CRN), $0.035 \pm 0.001$ in the baseline (GCRN+X), and $0.091 \pm 0.026$ in the ground truth.
From this viewpoint (both average and variance), our best model was closer to the ground truth than the baseline. 
Example \textcolor{black}{interpretable} results of our method are shown in Fig. \ref{fig:boid}.
Our best model (TG-CRN) shows better counterfactual covariate prediction with intervention than the baseline (GCRN$+$X).
Moreover, the counterfactual prediction of the outcome in our model had variation among various intervention times, whereas that in the baseline did not.
We consider that these were important properties in our problem, and improved prediction of the potential outcome was left for future work. 
\textcolor{black}{We expect that our approach can estimate the effect of an experimenter’s interventions on multi-animal behaviors even if they did not perform the experiment in missing conditions. Our approach is expected to improve the efficiency of experimental procedures for observing desired movements.}

\subsection{Real-world basketball dataset}
\label{ssec:basketball}
Finally, we examined the applicability of our methods to a real-world basketball dataset from the NBA.
Data acquisition was based on the contract between the league (NBA) and the company (STATS LLC.), not between the players and us. They are top-level players and then the data was not anonymized. The company was licensed to acquire this data, and it was guaranteed that the use of the data would not infringe on any rights of players or teams.
In this study, we used attack sequences from 630 games from the 2015/2016 NBA season (\url{ https://www.stats.com/data-science/}), which contained the trajectories of 10 players and the ball. 
We extracted 47,467 attacks (i.e., offensive plays) as samples, which were subsampled at 5 Hz.
We separated the dataset into 34,696 pre-training (458 games for training the following classifier of effective attack), 11460 training (154 games, 1/10 of that is used as validation), and 1,305 test samples (18 games) in chronological order.
Since scoring predictions are difficult in general (e.g., \cite{Fujii17,Fujii18}), 
we define the attack effectiveness as the outcome by predicting whether the attack is effective or not in the future.
This is because evaluating team movements based on scores alone may not provide a holistic view, due to factors such as the shooting skills of individual players.
In addition, we predict the attack effectiveness at the next time stamp using the pre-training samples and a logistic regression. 
The details are provided in Appendix \ref{app:nba}.

We verified our methods using factual data and provided insights using counterfactual prediction.
In model verification using factual data, we set $T=95$ and $T_b = 85$.
For counterfactual predictions, we set $T=105$ and $T_b = 95$ (at the end of attacks, i.e., a shot or turnover) and predicted all combinations of the counterfactual timing during the intervention period $T_i = 95,\ldots,98$. 
As the local covariates $x^l_t$, we used the position and velocity of all agents including the ball.
As the global covariates $x^g_t$, we used the ball player's information and areas (see also Appendix \ref{app:nba}), distances from the nearest defender (about the ball player and other attackers), successful shot probabilities of all attackers, and game and shot clock.
The theory-based function $f^x_{theory}$ and $f^y_{theory}$ computed 
$\hat{x}^g_{t+1}$ and $\hat{y}_{t+1}$ in rule-based manners.
In addition, to perform long-term prediction, $f^x_{theory}$ also computed the ball and the defender nearest to the next ball player during a counterfactual pass in rule-based manners.

\begin{figure}[h]
\centering
\includegraphics[width=0.45\textwidth]{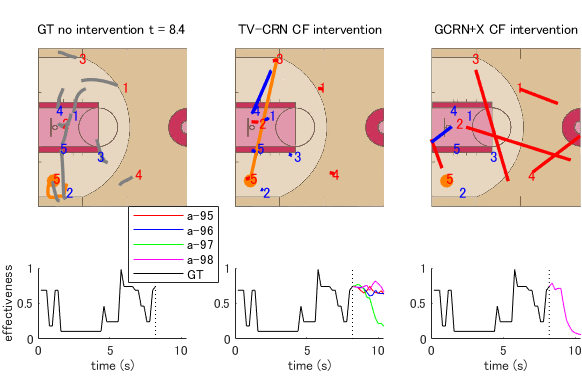}
\caption[]{Example NBA results of our method. The configurations are the same as Fig. \ref{fig:boid}. 
(Top) Red and blue numbers, gray line, and orange circle and line indicate an attacker, a defender, players' historical trajectories, and the ball, respectively.
The positions of the numbers are at the end of the factual data (shot), which is shown as the break line in lower plots. 
In the CF intervention subplots, colored trajectories indicate counterfactual predictions.
The actual red player \#5 shot (left) but in the counterfactual prediction (middle and right \textcolor{black}{columns}), the player tried to pass to a teammate. In the middle top (our method), the player successfully passes to the teammate red \#3, but in the left top (baseline), the player’s pass failed. 
\textcolor{black}{(Bottom) 
outcome time series (attack effectiveness) are shown. 
The ``a'' in the lower caption is the intervention times as shown in Fig. \ref{fig:boid}. For example, ``$a = 95$'' means the case of intervention at the 95th frame (9.5 s).}
\textcolor{black}{The video is given in the above GitHub page.}
}
\label{fig:nba}
\end{figure}

Our verification results using factual data are shown in Table \ref{tab:carla} right.
We examined the counterfactual pass effect $\hat{\tau}^{CFP} = \max_{t'\in T_i}(\max_{t=[T_b,T]}(\hat{y}^{(i,t')}_{T+1}))-y^{(i)}_{T_b+1}$ in addition to the outcome and covariate prediction errors. 
Our methods show better prediction performances in the factual covariates than the fully data-driven baselines. In the factual outcomes, our methods outperformed the most appropriate baseline (GCFN$+$X) and RNN, but only the method without GNN shows competitive performance with DSW and GCRN.  
One of the possible reasons may be the difficulty in modeling the relationship between the future outcome and current covariates, 
whereas GNN worked well in only the covariate prediction as the previous work \cite{Yeh19}.
\textcolor{black}{As additional analysis, we indicate endpoint errors and distributions of long-term covariate prediction for basketball data in Appendix \ref{app:analysis_basket}. Again, the strength of our method is to model the covariates at the next timestep for interpretability of the model. Results in Appendix \ref{app:analysis_basket} 
that the tendency of the endpoint prediction error for all models is similar to the mean prediction error, in which our approach (TV-CRN and TGV-CRN) show the best performance. In addition, the distribution of the player velocity in the figure of Appendix \ref{app:analysis_basket} 
shows that our approach without theory-based computation had a wider distribution like ground truth than other models, and those with theory-based computation had less zero-velocity bins like ground truth than other models. Although our approach did not completely model the player’s velocity, we show that they were partially effective for covariate prediction using each component.}
In $\hat{\tau}^{CFP}$, since all models show positive values, it suggests that there may be a more promising shot opportunity by an extra pass in the shot situation. 
Figure \ref{fig:nba} shows realistic counterfactual prediction from our model 
\textcolor{black}{to demonstrate interpretable results.}
Our model completed the counterfactual pass and the nearest defender chase, but the baseline model failed and predicted unrealistic behaviors 
(e.g., the attackers moved toward the outside of the court).
\textcolor{black}{In practice, our approach can estimate the effect of the selection of passes in basketball shot scenarios. We expect that our approach can evaluate the decision-making skills of players in a competitive game.}
Again, we emphasize these important properties in our problem. 

\section{Conclusions}
\label{sec:conclusion}
In this paper, we proposed an interpretable counterfactual recurrent network in multiagent systems to estimate the effect of the intervention.
Using synthetic CARLA and Boid datasets, we showed that our model achieved lower errors in estimating counterfactual covariates and the most effective treatment timing than the baselines.
Furthermore, using real basketball data, our model performed realistic counterfactual prediction.
We consider a general ITE framework for various domains, but the experimental results show that for each domain the effective modeling was different. 
Possible future research directions are to realize better modeling of the future outcomes, 
and to apply our approach to other multiagent domains such as animals and pedestrians using domain knowledge.

\section*{Acknowledgments}
This work was supported by JSPS KAKENHI (Japan Society for the Promotion of Science, Grant Numbers 20H04075, 21H04892, and 21H05300), JST PRESTO (Japan Science and Technology Agency, Precursory Research for Embryonic Science and Technology, Grant Number JPMJPR20CA), and JST CREST (Core Research for Evolutional Science and Technology, Grant Number JPMJCR1913).

\renewcommand{\thesection}{\Alph{section}}
\section*{Appendix}
\section{A proof of Theorem \ref{thm:identification}}
\label{app:proof}
\begin{proof}
Under the aforementioned assumptions in the main text, we can prove the identification of ITE:
\eq{\tau_t &= \mathbb{E}_y[ y_{1,t+1}-y_{0,t+1}|x_t,\mathcal{H}_t]\label{eq:proof1}\\
&=\mathbb{E}_z[\mathbb{E}_y[y_{1,t+1}-y_{0,t+1}|x_t,z_t,\mathcal{H}_t]|x_t,\mathcal{H}_t]\label{eq:proof2}\\
&=\mathbb{E}_z[\mathbb{E}_y[y_{1,t+1}-y_{0,t+1}|z_t]|x_t,\mathcal{H}_t]\label{eq:proof3}\\
&=\mathbb{E}_z[\mathbb{E}_y[y_{1,t+1}|z_t,a_t=1] - \mathbb{E}_y[y_{0,t+1}|z_t,a_t=0]|x_t,\mathcal{H}_t]\label{eq:proof4}\\
&=\mathbb{E}_z[\mathbb{E}_y[y_{F,t+1}|z_t,a_t=1] - \mathbb{E}_y[y_{F,t+1}|z_t,a_t=0]|x_t,\mathcal{H}_t],\label{eq:proof5}
}
where $\tau_t = \tau(x_t,\mathcal{H}_t)$, $y_{F,t+1}$ is a factual outcome, and we drop the instance index $(i)$ for simplification. 
Eq. (\ref{eq:proof1}) is the definition of ITE in our setting, Eq. (\ref{eq:proof2}) is a straightforward expectation over $p(z_t|x_t,\mathcal{H}_t)$ , and Eq. (\ref{eq:proof3}) be inferred from the structure of the causal graph shown in Fig. \ref{fig:causalgraph}.
Eq. (\ref{eq:proof4}) is based on the assumption that $z_t$ contains all the hidden confounders, as well as the positivity assumption, and Eq. (\ref{eq:proof5}) can be inferred from the consistency assumption. 
Thus, if our framework can correctly model $p(z_t|x_t,\mathcal{H}_t)$ and $p(y_t|z_t,a_t)$, then the ITEs can be identified under the causal graph in Fig. \ref{fig:causalgraph}.
\end{proof}

\section{Common training setup}
\label{app:common}
\subsection{Model training and computation}
\label{app:ourtraining}
The codes and data we used are provided at \url{https://github.com/keisuke198619/TGV-CRN}.
This experiment was performed on an Intel(R) Xeon(R) CPU E5-2699 v4 ($2.20$ GHz $\times$ 16) with GeForce TITAN X pascal GPU.
For the training of the proposed and baseline models, we used the Adam optimizer \cite{Kingma15} with an initial learning rate of $0.0001$ and $20$ training epochs.
We set the batchsize to 256.
For the hyper-parameters in the loss function, we set $\alpha = 0.1, \lambda = 0.1$ for all datasets and  $\gamma = 0.1$ in the Boid and CARLA dataset and  $\gamma = 1$ in the NBA experiment because the latter dataset was more difficult to predict than the Boid and CARLA experiments.

\subsection{Baseline models implementation}
\label{app:baselines}
We compared the performances of our methods to infer ITE with those in the following baselines: a simple RNN using GRU \cite{Cho14},
deep sequential weighting (DSW) \cite{liu2020estimating}, dynamic networked observational data deconfounder (but for clarity, we change the name into GCRN: graph counterfactual recurrent network) \cite{ma2021deconfounding}.

\noindent {\bf{RNN}.} This approach is based on GRU \cite{Cho14}. This model predicts the input (covariates) at the next time stamp, the potential outcome, and the probability of receiving treatment.
We also model the hidden confounder as the hidden state of GRU, but do not learn the representation to reduce the confounding bias.  

\noindent {\bf{DSW}} \cite{liu2020estimating}. Compared with the original model, we modified it to our setting (e.g., multiagent, time-varying outcome, and long-term prediction), and to fairly compare with our model, we removed the attention module. 

\noindent {\bf{GCRN}} \cite{ma2021deconfounding}. Similarly, compared with the original model, we modified it to our setting (e.g., multiagent and long-term prediction) based on DSW, and to fairly compare with our model, we removed the attention module.

\section{Sensitivity analysis in hyperparameters}
\label{app:sensitivity}

We performed the sensitivity analysis in hyperparameters using the CARLA dataset. 
The hyperparameters are presented in Eq. (\ref{eq:loss}).
Results in Table \ref{tab:sensitivity} shows the existence of the trade-off between the prediction performances of the outcome and covariates ($\gamma$).  

\begin{table}[ht!]
\centering
\scalebox{0.75}{
\begin{tabular}{l|cccc}
\Xhline{3\arrayrulewidth} 
& \me{1}{$L_{Outcome}$ } & \me{1}{$\sqrt{\epsilon^t_{PEHE}}$} &  \me{1}{$\epsilon^t_{ATE}$} & \me{1}{$L_{Covariates}$} \\
\hline
$\alpha, \gamma, \lambda = 0.1$ (default) & 0.041 $\pm$ 0.003 &  0.020 $\pm$ 0.002 & 0.008 $\pm$ 0.001 & 0.098 $\pm$ 0.0004 \\
\hline
$\alpha = 1.0, \gamma, \lambda = 0.1$ & 0.040 $\pm$ 0.003 &  0.020 $\pm$ 0.002 & 0.009 $\pm$ 0.001 & 0.102 $\pm$ 0.0004 
\\$\alpha = 0.01, \gamma, \lambda = 0.1$ & 0.041 $\pm$ 0.003 &  0.019 $\pm$ 0.002 & 0.007 $\pm$ 0.001 & 0.097 $\pm$ 0.0004 
\\$\gamma = 1.0, \alpha, \lambda = 0.1$ & 0.041 $\pm$ 0.003 &  0.019 $\pm$ 0.001 & 0.005 $\pm$ 0.001 & 0.086 $\pm$ 0.0004\\
$\gamma = 0.01, \alpha, \lambda = 0.1$ &  0.031 $\pm$ 0.002 &  0.020 $\pm$ 0.002 & 0.009 $\pm$ 0.001 & 0.146 $\pm$ 0.0005  
\\$\lambda = 1.0, \alpha, \gamma = 0.1$ & 0.062 $\pm$ 0.005 &  0.019 $\pm$ 0.002 & 0.007 $\pm$ 0.001 & 0.121 $\pm$ 0.0005  
\\$\lambda = 0.01, \alpha, \gamma = 0.1$ &  0.041 $\pm$ 0.003 &  0.020 $\pm$ 0.002 & 0.008 $\pm$ 0.001 & 0.098 $\pm$ 0.0004
\\
\Xhline{3\arrayrulewidth} 
\end{tabular}
}
\caption{\label{tab:sensitivity} Sensitivity analysis on the carla dataset.}
\end{table}

\vspace{-10pt}
\section{Boid dataset}
\label{app:boid}
The schooling model we used in this study was a unit-vector-based (rule-based) model \cite{Couzin02}, which accounts for the relative positions and direction vectors of neighboring fish agents, such that each fish tends to align its own direction vector with those of its neighbors. 
In this model, $20$ agents (length: 0.5 m) are described by a two-dimensional vector with a constant velocity (1 m/s) in a boundary square (30 $\times$ 30 m) as follows: 
${r}^k=\left({x_i}~{y_i}\right)^T$ and ${v}^k_t= \|v^k\|_2d_k$, where $x_i$ and $y_i$ are two-dimensional Cartesian coordinates, ${v}^k$ is a velocity vector, $\|\cdot\|_2$ is the Euclidean norm, and $d_k$ is an unit directional vector for agent $i$.

At each timestep, a member will change direction according to the positions of all other members. The space around an individual is divided into three zones where each modifying the unit vector of the velocity (for the zones, see the main text).
Let $\lambda_r$, $\lambda_o$, and $\lambda_a$ be the numbers in the zones of repulsion, orientation and attraction respectively. For $\lambda_r \neq 0$, the unit vector of an individual at the next timestep is given by:
\begin{equation} 
\label{eq:boid1}
d_k(t+1 , \lambda_r \neq 0 )=-\left(\frac{1}{\lambda_r-1}\sum_{j\neq k}^{\lambda_r}\frac{r^{kj}_t}{\|r^{kj}_t\|_2}\right),
\end{equation} 
where $r^{kj}={r}_j-{r}_i$.
The velocity vector points away from neighbors within this zone to prevent collisions. This zone is given the highest priority; if and only if $\lambda_r = 0$, the remaining zones are considered. 
The unit vector in this case is given by:
\begin{equation} 
\label{eq:boid2}
d_k(t+1 , {\lambda}_r=0)=\frac{1}{2}\left(\frac{1}{\lambda_o}\sum_{j=1}^{\lambda_o} d_j(t)+\frac{1}{\lambda_a-1}\sum_{j\neq k}^{{\lambda}_a}\frac{r^{kj}_t}{\|r^{kj}_t\|_2}\right).
\end{equation} 
The first term corresponds to the orientation zone while the second term corresponds to the attraction zone. The above equation contains a factor of $1/2$ which normalizes the unit vector in the case where both zones have non-zero neighbors. If no agents are found near any zone, the individual maintains a constant velocity at each timestep.

In addition to the above, we constrain the angle by which a member can change its unit vector at each timestep to a maximum of $\beta = 30$ deg. This condition was imposed to facilitate rigid body dynamics. Since we assumed point-like members, all information about the physical dimensions of the actual fish is lost, which leaves the unit vector free to rotate at any angle. In reality, however, the conservation of angular momentum will limit the ability of the fish to turn angle $\theta$ as follows:
\begin{equation} 
\label{eq:boid3}
  d_k\left(t+1 \right)\cdot d_k(t) = 
  \begin{cases}
   \cos(\beta ) & \text{if $\theta >\beta$} \\
   \cos\left(\theta \right) & \text{otherwise}.
  \end{cases}
\end{equation} 
If the above condition is not satisfied, the angle of the desired direction at the next timestep is rescaled to $\theta = \beta$. In this way, any un-physical behavior such as having a 180$^\circ$ rotation of the velocity vector in a single timestep is prevented.

In the simulation, the ground truth of $\tau_T$ was $0.091 \pm 0.026$, which indicates the intervention increased angular velocities (see also the main text).


\section{CARLA dataset}
\label{app:carla}
We used the CARLA simulator \cite{dosovitskiy2017carla} (ver. 0.9.8).
The CARLA environment contains dynamic obstacles (e.g., pedestrians and cars) that interact with the ego car. For generating data, we performed simulation through various towns, starting points, and obstacle positions, which were subsampled at 2 Hz.
The obstacles' positions and starting points of the ego car were randomly selected from the possible locations on the map for each run.
Approximately 20-120 obstacles were placed on each map.
Using the data collected in CARLA, the same driving conditions were reproduced in ROS (robot operating system) \cite{quigley2009ros}, and Autoware was used to make the vehicle drive the same route autonomously.
Here we consider two types of autonomous vehicles: partial observation with Autoware \cite{kato2015open} (ver.1.12.0) and full observation types with the pre-installed CARLA simulator. 
The partial observation model often stops when dangerous situations for safety. 
If an obstacle enters the deceleration or stopping range, the vehicle decelerates or stops in front of the obstacle.
The deceleration range was set to be wider than usual for safety.
The full observation model intervenes to accelerate the stopped ego car after the safety confirmation based on the speed information at the data collection.

We set $T=60, T_b=40$, and $T_i=T_b+1,\ldots,T_b+10$.
We evaluated and predicted the safe driving distance of the ego car from the starting point while driving without collision with any obstacle (for details, see the main text). 
In the simulation, the ground truth of $\tau_T$ was $0.013 \pm 0.001$, which indicates the intervention increased safe driving distance. 
\section{NBA dataset}
\label{app:nba}
Here, we describe the details of the computation using the NBA dataset.
Dataset description is given in the main text.
We describe the computation of the attack effectiveness used in this study.
Then, we predict the effective attacks at the next time stamp using the pre-training samples and a logistic regression.  
%
%
%

%
From the aforementioned reasons in the main text, we compute an interpretable and simple indicator from available statistics (i.e., based on the frequency) to evaluate whether a player attempts a better shot, rather than based on the shot label or learning-based score prediction.
From available statistics, we focused on two basic factors for effective attacks at an individual player level: the shot zone on the court and the distance between a shooter and the nearest defender. 
These two factors have been considered to be important for basketball shot prediction \cite{Fujii16,Fujii17,Fujii18}.
In the NBA advanced stats (\url{https://www.nba.com/stats/players/shots-closest-defender/}), we can access probabilities of successful shots in each zone and distance for each player.
The shot zones are separated into four areas: the restricted area, in-the-paint, mid-range, and the 3-point area. 
The restricted area is defined as the area within a radius of 2.44 m from the ring (distance between the side of the rectangle and the ring) from the ring.
In-the-paint is defined as the area within a radius of 5.46 m (distance between the ring and the farthest vertex of the rectangle) from the ring.
The 3-point area is defined as the outside of the 3-point line. 
Mid-range is the remaining area. 
The shooter's distance from the nearest defender is categorized into four ranges: $0-2$ feet, $2-4$ feet, $4-6$ feet, and $6+$ feet.

We define the attack effectiveness using the following criteria:
\vspace{-0pt}
\begin{quote}
\begin{itemize}
\item The shooter's position in the restricted area is effective at any distance because the defender often exists near the shooter.
\item The shooter's position in the paint and mid-range is effective at 6 feet or further (this range is regarded as ``open'' in the NBA advanced stats).
\item The shooter's position in the 3-point area is effective when a player with a probability of 0.35 and hits with 6 feet or more (because some players do not shoot tactically).
\end{itemize}
\end{quote}
\vspace{-0pt}
Based on the statistics before the game (e.g., for the pre-training data, we used the 2014/2015 season statistics) and the tracking data, 
we computed the probabilities of successful shots for each zone and the distances for each player. 
We computed the probability of the player who attempted shots less than 10 times as the probability of the same position player (i.e., guard, forward, center, guard/forward, and forward/center based on the registration information from the NBA 2014/2015 season).
It should be noted that there are some strategies of a good shot in basketball that differ depending on the court location and context, for example, 2 pointers and 3 pointers.
Note that, unfortunately, we can access those for only two areas (2-point and 3-point areas) with four distance categories, thus we computed the shot success probability at the restricted area, in-the-paint, and mid-range using that at the 2-point area.
%
Based on the above definitions, for pre-training data, there were 13,681 shot successes, 15,443 shot failures, 5,572 turnovers,
16,976 effective attacks, and 17,720 ineffective attacks.
For training data, there were 4,677 shot successes, 5,092 shot failures, 1,691 turnovers, 5,602 effective attacks, and 5,858 ineffective attacks.
For test data, there were 517 shot successes, 577 shot failures, 211 turnovers, 664 effective attacks, and 641 ineffective attacks.
The probabilities of scoring, given the attack was effective and ineffective, were 0.463/0.415/0.417 and 0.328/0.402/0.374 for pre-training, training, and test data, respectively.
We confirmed that the effective attack had a higher probability of a successful shot than the ineffective attack. 

After the computation of the attack effectiveness, we predict the effective attacks at the next time stamp using the pre-training samples and a logistic regression.  
In the soccer domain, there has been some work to evaluate players and teams (e.g., \cite{Decroos19,Toda2021}) based on the prediction model (i.e., classifier) on the assumption that a good play is a play that will bring good outcomes (e.g., score and ball recovery) in the future. 
These approaches can transform a discrete value (i.e., an outcome) into a continuous value (e.g., a probability).
According to these papers, we also predict the effective attack at the next time stamp.
To verify the prediction accuracy, we compared the accuracy of the logistic regression, LightGBM \cite{ke2017lightgbm}, which is a popular classifier using a highly efficient gradient boosting decision tree, and prediction with all effective attacks.
For the training data, the accuracies of logistic regression, LightGBM, and the prediction with all the same labels were 0.838, 0.838, and 0.731, respectively.
For the test data, these were 0.679, 0.676, and 0.568, respectively. 
We confirmed that the logistic regression had competitive performance compared with LightGBM, and higher accuracy than the prediction with all the same labels, even maintaining higher interpretability.

\section{\textcolor{black}{Additional analysis in basketball dataset}}
\label{app:analysis_basket}
\begin{figure}[h]
\centering
\includegraphics[width=0.5\textwidth]{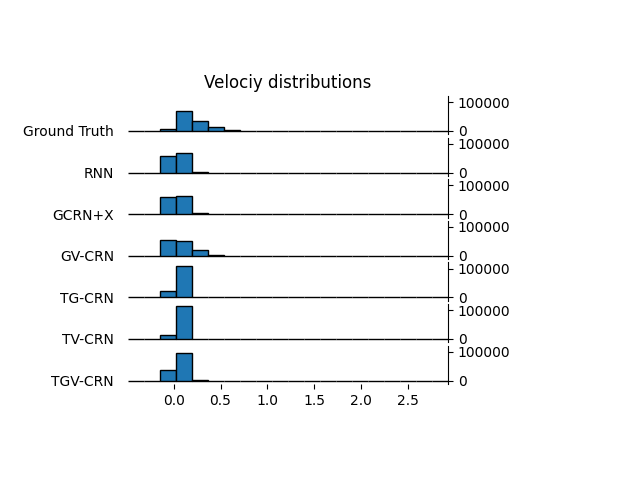}
\caption[]{\textcolor{black}{Histograms of player velocity in NBA dataset.}
}
\label{fig:histogram_nba}
\end{figure}

\textcolor{black}{As additional analysis, we indicate endpoint errors and distributions of long-term covariate prediction for basketball data.
Results in Table \ref{tab:nba_end} that the tendency of the endpoint prediction error for all models is similar to the mean prediction error, in which our approach (TV-CRN and TGV-CRN) show the best performance. In addition, the distribution of the player velocity in Fig. \ref{fig:histogram_nba} shows that our approach without theory-based computation had a wider distribution like ground truth than other models, and those with theory-based computation had less zero-velocity bins like ground truth than other models. }
\begin{table}[ht!]
\centering
\scalebox{1}{
\begin{tabular}{l|c}
\Xhline{3\arrayrulewidth} 
 & \me{1}{$L_{Covariates}$ at endpoint} \\
\hline
RNN & 1.637 $\pm$ 0.0096
\\DSW \cite{liu2020estimating}& ---
\\GCRN \cite{ma2021deconfounding} & ---
\\GCRN$+$X & 2.266 $\pm$ 0.0138\\
\hline
GV-CRN & 1.937 $\pm$ 0.0116
\\TG-CRN & \textbf{1.482} $\pm$ \textbf{0.0090}
\\TV-CRN & 1.566 $\pm$ 0.0094
\\ TGV-CRN (full) & \textbf{ 1.510} $\pm$ \textbf{0.0093}
\\
\Xhline{3\arrayrulewidth} 
\end{tabular}
}
\vspace{3pt}
\caption{\label{tab:nba_end} \textcolor{black}{Endpoint prediction error on NBA dataset.}}
\end{table}

\section{\textcolor{black}{Convergence in the learning of our models}}
\label{app:converge}
\begin{figure}[h]
\centering
\includegraphics[width=0.5\textwidth]{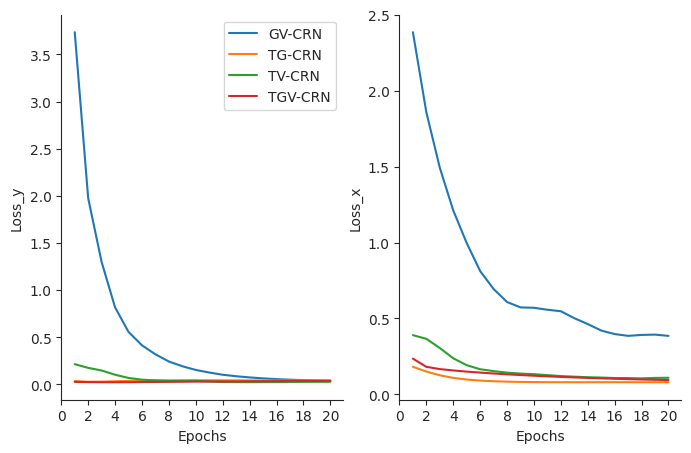}
\caption[]{\textcolor{black}{The learning curve of our model training in CARLA dataset. Left and right subfigures indicate losses in outcome and covariate prediction, respectively. }
}
\label{fig:carla_learning}
\end{figure}

\begin{figure}[h]
\centering
\includegraphics[width=0.5\textwidth]{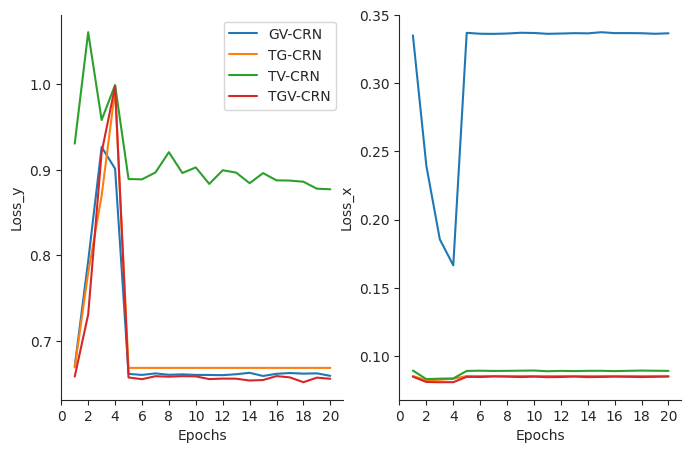}
\caption[]{\textcolor{black}{The learning curve of our model training in Boid dataset.}
}
\label{fig:boid_learning}
\end{figure}

\begin{figure}[h]
\centering
\includegraphics[width=0.5\textwidth]{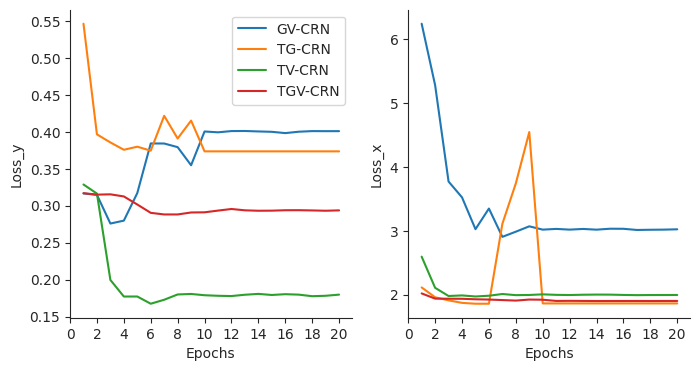}
\caption[]{\textcolor{black}{The learning curve of our model training in NBA dataset.}
}
\label{fig:nba_learning}
\end{figure}

\textcolor{black}{We illustrate the change in the validation losses of our models over the course of training epochs as shown in Figs. \ref{fig:carla_learning}, \ref{fig:boid_learning}, and \ref{fig:nba_learning}.
Compared with CARLA dataset results, which show better outcome and covariate prediction results than other datasets, those in Boid and NBA datasets sometimes show unstable learning curves, but most of the models finally show convergence in all datasets. 
To perform fair comparisons among the models, we trained all models with 20 epochs. }

%
\bibliographystyle{IEEEtran}
\bibliography{main}

\clearpage

\ifCLASSOPTIONcaptionsoff
  \newpage
\fi

\end{document}